\documentclass[numsec,webpdf,modern,large,namedate]{{oup-authoring-template}}
\graphicspath{{figs/}}

\theoremstyle{thmstyleone}%
\theoremstyle{thmstyletwo}%
\theoremstyle{thmstylethree}%

\usepackage{subfigure}
\usepackage{pifont}
\usepackage{algorithm}
\usepackage{algpseudocode}
\usepackage{xcolor}
\usepackage{booktabs}

\makeatletter
\renewcommand{\@thesubfigure}{\footnotesize(\alph{subfigure})}
\makeatother
\begin{document}

\journaltitle{Bioinformatics}
\DOI{\url{https://doi.org/10.1093/bioinformatics/btag575}}
\copyrightyear{2026}
\pubyear{2026}
\access{Published: 5 August 2026}
\appnotes{Original Paper}

\firstpage{1}


\title[ConfRetro]{ConfRetro: A 3D-aware Template-free Method for Enhancing Retrosynthesis via Molecular Conformer Information}

\author[1,2]{Jiaxi Zhuang\ORCID{0009-0009-9723-9635}}
\author[1,2]{Yu Zhang\ORCID{0009-0003-8099-4710}}
\author[1,3,4,$\ast$]{Ying Qian\ORCID{0000-0003-4961-5842}}
\author[1,3,4,$\ast$]{Aimin Zhou\ORCID{0000-0002-4768-5946}}

\authormark{Jiaxi Zhuang et al.}
\address[1]{Shanghai Frontiers Science Center of Molecule Intelligent Syntheses, Shanghai 200062, China}
\address[2]{School of Computer Science and Technology, East China Normal University, Shanghai 200062, China}
\address[3]{Shanghai Institute of Artificial Intelligence for Education, East China Normal University, Shanghai 200062, China}
\address[4]{Lab of Al for Educalion, East China Normal University, Shanghai 200062, China}

\corresp[$\ast$]{Corresponding author. \href{email:email-id.com}{yqian@cs.ecnu.edu.cn}; \href{email:email-id.com}{amzhou@cs.ecnu.edu.cn}}

\received{14}{1}{2026}
\revised{24}{6}{2026}
\accepted{11}{7}{2026}

\abstract{
\textbf{Motivation:} Retrosynthesis plays a crucial role in organic synthesis and drug discovery, focusing on identifying a set of reactants capable of synthesizing a target product molecule. Although the existing approaches have shown promising results, they do not fully exploit 3D conformer information and molecular spatial structure, which can hinder stereochemically consistent and chemically plausible predictions.
\\ \textbf{Results:} To tackle this problem, we propose ConfRetro, a Transformer-based template-free method that integrates molecular conformer information and spatial structure. We devise an Atom-align Fusion module to combine 3D positional information at model input stage, ensuring alignment between atom tokens and corresponding 3D representations. Furthermore, we design a Distance-weighted Attention mechanism to guide self-attention, constraining receptive field of model and emphasizing chemically relevant atom pairs in 3D space. Experiments conducted on the USPTO-50K and USPTO-FULL datasets demonstrate that ConfRetro significantly outperforms existing template-free approaches, achieving a new state-of-the-art performance. Case studies further highlight its capability to predict accurate and chemically plausible reactants, even for target molecules with intricate structures. Moreover, when plugged into a standard retrosynthetic search, ConfRetro recovers feasible synthetic routes for multiple representative drug molecules (\emph{e.g.}, Camptothecin).
\\ \textbf{Availability and implementation:} The ConfRetro is available at \url{https://github.com/Jesse-zjx/ConfRetro}. Archival snapshot with \url {https://doi.org/10.5281/zenodo.20785018}.
}

\maketitle 

\section{Introduction}

Retrosynthesis, initially formulated by \citep{retrofirst}, stands as a fundamental problem within organic synthesis, aiming to discover and predict reactants given product molecules. This task presents a large chemical search space and myriad molecular combinations, posing significant challenges. Traditionally, researchers relied heavily on extensive knowledge of organic chemistry and reaction mechanisms, often resulting in inefficiency and heavy dependency on expertise. In recent years, with the significant impact of data-driven deep learning methods across various fields, computer-aided retrosynthesis has gained increasing attention, especially the template-free methods \citep{recent}. 

Existing methods can be broadly categorized into three types based on reliance on template prior: template-based, semi-template, and template-free. Template-based methods \citep{localretro,retrosim,gln} view retrosynthesis as a template retrieval task and subsequently employ the retrieved templates (specific chemical changes) to transform the input product molecule into reactants using cheminformatics tools like {\tt RDKit} \citep{rdkit}. While these methods offer strong interpretability, their reliance on pre-defined templates limits their coverage and scalability, rendering them insufficient for real-world scenarios. Semi-template methods \citep{g2gs,xpert} follow a two-stage pipeline: (i) reaction center identification, where product-reactant aligned atom mapping is used to predict disconnection bonds and convert products into incomplete molecules (synthons), and (ii) synthon completion, where generative models are employed to recover full reactants from synthons. Representative works include RetroXpert \citep{xpert}, which employs an Edge-enhanced Graph Attention Network (EGAT) for bond disconnection followed by a seq2seq model for synthon completion, and G2Gs \citep{g2gs}, which completes synthons in graph format. Although semi-template methods remove the strict dependency on pre-defined templates, the two-stage optimization process can cause suboptimal performance and weak generalization. Template-free methods \citep{scrop,augtrans,get} treat retrosynthesis prediction as a sequence-to-sequence machine translation task, typically using SMILES \citep{smiles} to represent molecules. Transformer-based approaches \citep{augtrans,retroformer} have further improved scalability and implicit learning of chemical rules, while data augmentation and permutation invariance strategies \citep{graph2smiles,gta} enhance robustness. Nevertheless, most existing approaches primarily rely on 1D SMILES or 2D graphs, and only weakly encode explicit 3D geometric cues. While such representations have been effective on large reaction corpora, the lack of 3D-aware inductive biases can make it harder to model stereochemistry-sensitive transformations (e.g., chirality and spatial constraints) and may reduce robustness on structurally complex targets \citep{r3d,m3d}. In particular, models trained purely from 1D/2D inputs may produce reactant predictions with stereochemical inconsistencies or implausible configurations for molecules with rich 3D structure, such as polychiral or heteroaromatic compounds \citep{deeplinker}.

\textbf{Problem.} In previous studies, the typical representation of molecules involves 1D sequences (\emph{i.e.}, SMILES) and 2D graphs, where explicit 3D conformational cues are largely absent. This omission may hinder accurate perception of stereochemistry and reaction feasibility. Incorporating 3D conformers into retrosynthesis prediction poses two challenges: (i) integrating 3D features while maintaining semantic alignment between atom tokens and corresponding 3D representations; and (ii) constraining the receptive field of attention mechanisms to better reflect spatial molecular structure.

In this work, we propose ConfRetro, a novel end-to-end, template-free retrosynthesis model that integrates 3D conformer information into a Transformer framework. We design two key modules: \emph{Atom-align Fusion}, which integrates 1D and 3D representations while preserving atom-level alignment, and \emph{Distance-weighted Attention}, which redistributes attention weights based on inter-atomic spatial distances. Moreover, we employ data augmentation strategies \citep{augtrans,rsmiles} such as SMILES permutation and atom alignment, as well as attention guidance loss \citep{retroformer} to further enhance robustness and training efficiency. 

Our contributions can be summarized as follows:
\begin{itemize}
    \item We propose ConfRetro, an end-to-end template-free retrosynthesis model that incorporates 3D conformer information to improve stereochemistry-aware prediction and robustness on structurally complex molecules.
    \item We introduce two novel modules: i) \emph{Atom-align Fusion} to integrate SMILES and 3D representations while preserving their semantic alignment; ii) \emph{Distance-weighted Attention} to constrain receptive fields based on 3D distances between atoms.
    \item Extensive experiments on USPTO-50k \citep{uspto50k} and USPTO-FULL datasets demonstrate that ConfRetro achieves state-of-the-art performance among template-free methods, under both with and without reaction class settings, and is also competitive with template-based and semi-template-based approaches.
\end{itemize}

\section{Preliminary}

\subsection{Template-free Retrosynthesis} 
Molecules can be represented as sequences of tokens in SMILES format. Following the SMILES tokenization of \citep{moltrans} that separates each atom (\emph{i.e.}, {\tt O, C, N, Cl}) and non-atom tokens (\emph{i.e.}, {\tt -, =, \#}), we let $S=[s_1, s_2, \dots s_n]$ denote SMILES sequence with $n$ tokens. In template-free retrosynthesis, given the product SMILES $S_P$, the goal is to model the conditional distribution of the reactant SMILES $S_R$ with an auto-regressive sequence-to-sequence model:
\begin{equation}
p_\theta(S_R \mid S_P) = \prod_{t=1}^{|S_R|} p_\theta(r_t \mid r_{<t}, S_P),
\end{equation}
where $r_t$ denotes the $t$-th token in $S_R$ and $r_{<t}$ are the previously generated reactant tokens.

\subsection{Molecular Conformer and ComENet}
Molecule is a dynamic structure since atoms are in continual motion in 3D space. The local minima on the potential energy surface are called conformers. Generally, molecular conformer can be represented as $G_{\text{3D}}=(A, C)$, where $A=\{a_i\}_{i=1,\dots,n}$ is atoms list, $C=\{(x_i, y_i, z_i)\}_{i=1,\dots,n}$ is 3D coordinates list. To fulfill the global completeness of 3D conformer information of a molecule, ComENet \citep{comenet} encodes rotation angles and spherical coordinates within the 1-hop neighborhood of atoms via the message passing scheme \citep{mpnn}.

\begin{figure*}
  \centering
  \includegraphics[width=\linewidth]{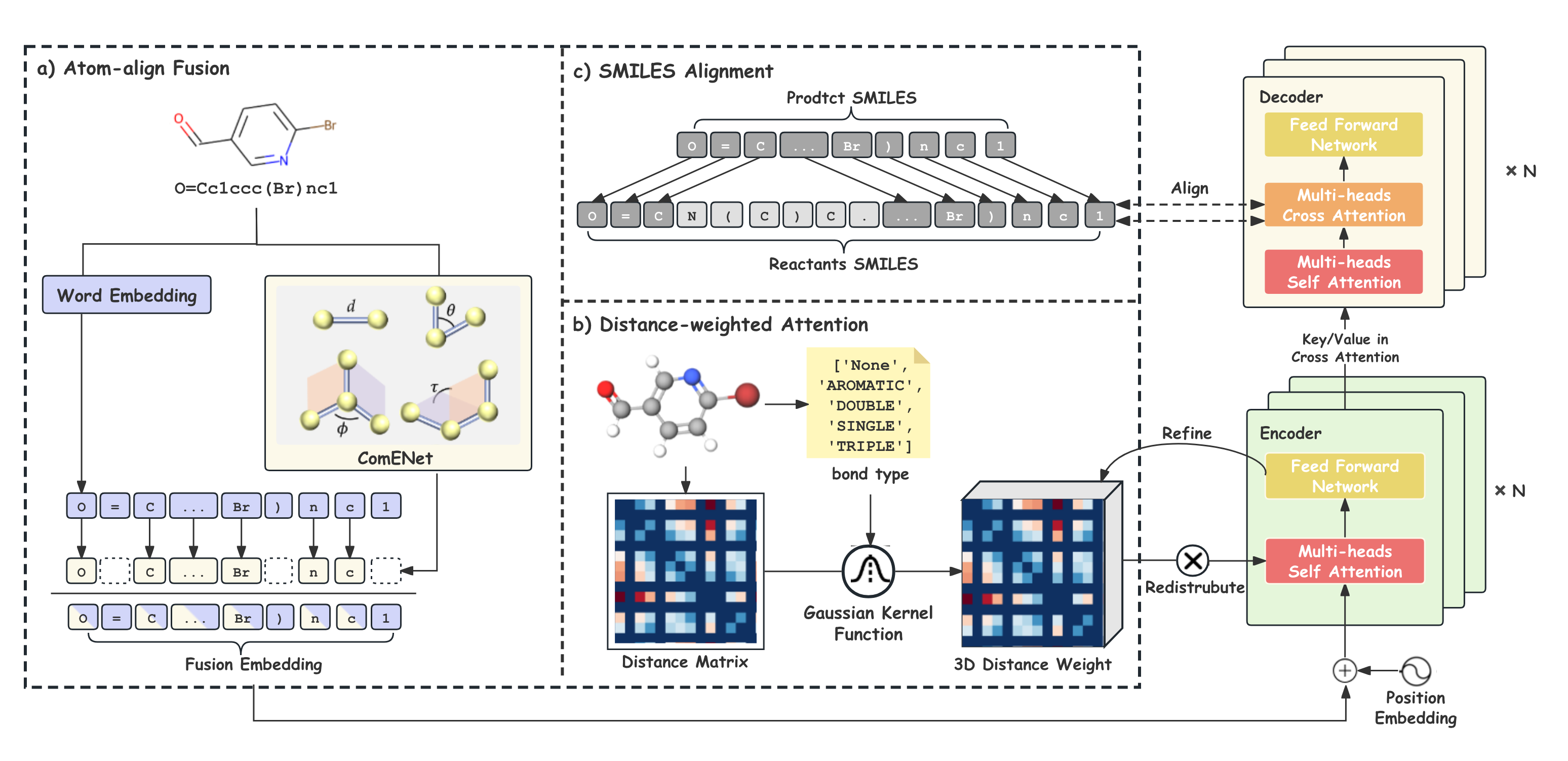}
  \caption{ Architecture overview of ConfRetro based on Transformer. a) SMILES and molecular conformer are processed through Word Embedding and ComENet to yield SMILES sequence embedding and 3D position embedding. Following the Atom-align integration, fusion embedding is then input into the Encoder. b) In Encoder, 3D Distance weight is obtained for redistribution of self-attention by Gaussian Basis Kernel function, while also refined layer-by-layer. c) In Decoder, SMILES Alignment is employed as guidance to align the cross-attention between the product and reactants.} 
  \label{fig1}
\end{figure*}

\section{Method}
In this section, we introduce ConfRetro, a novel transformer-based model that is capable of learning richer molecular representation through SMILES sequence and spatial conformer. An overview of the proposed method is shown in Figure~\ref{fig1}. 

Figure~\ref{fig1}a illustrates Atom-align Fusion, which incorporates both 1D SMILES token embedding and 3D position embedding to obtain a fusion embedding as input to the encoder-decoder model. By adopting this approach, the model is capable of integrating 3D positional information while preserving alignment (Section~\ref{sec4.1}). Figure~\ref{fig1}b illustrates Distance-weighted Attention, which directs self-attention to prioritize chemically relevant atoms within spatial structure (Section~\ref{sec4.2}). Figure~\ref{fig1}c illustrates SMILES Alignment strategy, which further bridges the mapping between product and reactant SMILES in cross-attention module of decoder (Section~\ref{sec4.3}). The overall training and inference process is conducted in an end-to-end manner, and it is a template-free method that does not rely on {\tt RDKit} for molecular editing.

\subsection{Atom-align Fusion} \label{sec4.1}
The most direct method of feature integration is concatenation \citep{get}, but this approach disrupts the alignment between 1D representations of atom tokens and their corresponding 3D representations. The challenge lies in how to integrate conformer information into the molecular representation while preserving alignment between them. 

To address this issue, we introduced the Atom-align Fusion module. The purpose is to obtain fusion embedding $F_{\text{3D}} \in \mathbb{R}^{M \times D}$ by associating SMILES token embedding $T \in \mathbb{R}^{M \times D}$ with 3D position embedding $P_{\text{3D}} \in \mathbb{R}^{N \times D}$, where $D$ is model dimension, $M$ is number of SMILES tokens, and $N$ is number of atoms.

Firstly, to obtain the complete 3D  molecular representation, we extracted 3D position embedding $P_{\text{3D}}$ from the molecular conformer using ComENet \citep{comenet}. ComENet proposes a novel message passing scheme \citep{mpnn} that operates within a 1-hop neighborhood with radial distance $d$, polar angle $\theta$, azimuthal angle $\phi$ and rotation angle $\tau$, and then builds message passing scheme as:
\begin{equation}
\mathbf{v}_i^{t+1}=g\left(\mathbf{v}_i^{t}, \sum_{j \in \mathcal{N}_i} f\left(\mathbf{v}_j, d_{i j}, \theta_{i j}, \phi_{i j}, \tau_{i j}\right)\right),
\label{equ1}
\end{equation}
where $\mathbf{v}_i^{t} \rightarrow \mathbf{v}_i^{t+1}$ denotes the update of atom (node) representation, $g$ and $f$ are implemented by neural networks or mathematical operations, $d_{i j}$, $\theta_{i j}$ and $\phi_{i j}$ specify the 1-hop local neighborhood, and $\tau_{i j}$ determines its orientation. We omit the final aggregation layer of ComENet to obtain the 3D representation corresponding to each atom $P_{\text{3D}}=\{\mathbf{v}_i\}_{i=1,2,\dots,N}$ rather than the entire molecular graph.

Subsequently, we input $P_{\text{3D}}$ along with the SMILES token embedding $T$ into the encoder. To facilitate a more comprehensive integration of SMILES and conformer, we align the embeddings $T$ and $P_{\text{3D}}$ by atom index in the same vector space. We first enlarge the length of $P_{\text{3D}}$ from $N$ (the number of atoms) to $M$ (the number of tokens) by padding zeros at the indices of non-atom tokens. At the model input stage, we blend $P_{\text{3D}}$ and $T$ to obtain the fusion embedding $F_{\text{3D}}$, allowing effective integration of 3D positional and sequential information at each atom position while preserving the alignment between them:

\begin{equation}
F_{\text{3D}} = \lambda_{1} \times \operatorname {pad}(P_{\text{3D}}) + \lambda_{2} \times T,
\label{equ3}
\end{equation}
where $\lambda_{1}$ and $\lambda_{2}$ are trainable parameters that adaptively adjust the weights of 3D position embedding $P_{\text{3D}}$ and SMILES token embedding $T$ during training. In addition, we still use Sinusoidal Position Embedding \citep{transformer} to encode the positions of tokens.

\subsection{Distance-weighted Attention} \label{sec4.2}
According to \citep{localattn,retroformer}, multi-head attention calculation in the vanilla Transformer is insufficient and lacks chemical knowledge. Simultaneously, 3D distances between atoms within a molecule reflect the spatial relation between any pair of atoms in physical chemistry, such as bond energy, stability, electron affinity and interatomic force. Incorporating 3D distances into the multi-head attention calculation process enables the model to reschedule attention accordingly, which not only alleviates the redundancy in multi-head attention but also enables distance-based attention redistribution during the attention calculation.

Similar to the method of partitioning attention heads in \citep{retroformer}, we divide the multi-head attention into normal attention heads and spatial attention heads. Normal attention heads follow the traditional transformer manner to learn sequence-wide information, whereas spatial attention heads employ 3D distances as an external attention weight to enable attention redistributed based on distances. The whole module can be formulated as two steps: 1) Construct 3D distance weight, 2) Attention Redistribute \& Weight Refine.

\subsubsection{Construct 3D Distance Weight.} First, we construct a distance matrix $\mathbf{D} \in \mathbb{R}^{N \times N}$ from molecular conformer as:
\begin{equation}
\mathbf{D}=\{\mathbf{d}_{i,j}\}_{i,j=1,2,\dots,N} \quad \mathbf{d}_{ij}=\|\mathbf{c}_{i}-\mathbf{c}_{j}\|_2 ,
\end{equation}
where $N$ is the number of atoms, $i,j$ are the indices of an atom pair, and $\mathbf{c}_{i}$ is the coordinate of the $i$-th atom. To explore high-dimensional space, we employ Gaussian Basis Function \citep{gbf} to transform the distance $\mathbf{d}_{ij}$ between each atom pair ($i$, $j$) into a higher dimension as:
\begin{equation}
\resizebox{.91\linewidth}{!}{$
            \displaystyle
\psi_{(i, j)}^k=
\frac{1}{\sqrt{2 \pi}\left|\sigma^k\right|} \exp \left(-\frac{1}{2}\left(\frac{\gamma_{(i, j)}\mathbf{d}_{ij}+\beta_{(i, j)}-\mu^k}{\left|\sigma^k\right|}\right)^2\right),
$}
\end{equation}
where $k$=\{1, \ldots, K\} and $K$ is the number of Gaussian kernels. $\mu^k$ and $\sigma^k$ represent the center and scaling factor of the $k$-th Gaussian basis kernel. $\gamma_{(i, j)}$ and $\beta_{(i, j)}$ are learnable scalars related to bond type (\emph{e.g.}, SINGLE, DOUBLE) of atom pair ($i$, $j$). To obtain the 3D Distance weight $\Phi_{(i, j)}$, we use two layers of Non-Linear Layer with the activation function Gaussian Error Linear Units ($\operatorname{GELU}$) \citep{gelu} to encode $\psi_{(i, j)}$ as:
\begin{equation}
\Phi_{(i, j)}= \mathbf{W_2}(\operatorname {GELU}(\mathbf{W_1}(\psi_{(i, j)}))),
\end{equation}
where $\psi_{(i, j)}=\operatorname{stack}([\psi_{(i, j)}^1;\dots;\psi_{(i, j)}^k])$, $\mathbf{W_1}, \mathbf{W_2} \in \mathbb{R}^{K \times K}$ are learnable parameters and 3D Distance weight $\Phi_{(i, j)} \in \mathbb{R}^{N \times N \times K}$.

\subsubsection{Attention Redistribute \& Weight Refine.} Second, we leverage the previously obtained 3D Distance weight $\Phi_{(i, j)}$ to guide the model in redistributing attention weights by the internal spatial structure, instead of directly calculating attention scores between each token like vanilla Transformer. We divide the attention heads into two halves: normal attention head and spatial attention head. The normal attention head still captures the context of the SMILES sequence like original attention computation. The spatial attention head, on the other hand, focuses more on the molecular internal structure, limiting the receptive field of each atom and considering the 3D distance with other atoms. For the $i$-th token of the $l$-th encoder layer, the roll-out form of spatial attention head is:
\begin{equation}
\begin{aligned}
x_{i,\text{spatial}}^{l+1} & =\sum_{j \in N(i)} \operatorname {softmax} \left(\Phi_{(i, j)}^l \odot \frac{q_i k_j^T}{\sqrt{d}}\right) v_j, \\
{\left[q_i, k_j, v_j\right] } & =\left[h_i^l \mathbf{W^Q}, h_j^l \mathbf{W^K}, h_j^l \mathbf{W^V}\right],
\end{aligned}
\end{equation}
where $\Phi_{(i, j)}^l$ is almost the same as 3D Distance weight $\Phi_{(i, j)}$ but padding 0 in non-atom token for element-wise multiplication $\odot$, $\mathbf{W^Q}$, $\mathbf{W^K}$ and $\mathbf{W^V}$ are projection matrices for query $q$, key $k$, value $v$ in self-attention. Computed representations of normal attention head and spatial attention head are concatenated along the hidden dimension and then processed through a linear layer, forming the updated context for the next layer $h^{l+1}$, formally we have:
\begin{equation}
\begin{aligned}
h^{l+1} =\mathbf{W_{Linear}}\left(\left[x^{l+1}_{\text {normal}} ; x^{l+1}_{\text{spatial}}\right]\right).
\end{aligned}
\end{equation}
In addition, we perform Weight Refine to update our 3D Distance weight layer by layer. The Weight Refine module is a fully connected (FC) layer that takes concatenation of updated contexts of atoms pair ($i$, $j$) to refine the 3D attention weight among receptive field, which is formulated as:
\begin{equation}
\Phi_{(i, j)}^{l+1} =\Phi_{(i, j)}^l+\operatorname{FC}\left(\left[h_i^{l+1} ; h_j^{l+1}\right]\right).
\end{equation}

\subsection{SMILES Alignment} \label{sec4.3}
During the occurrence of chemical reactions, the majority of atoms remain unchanged \citep{localretro,rsmiles}. We can utilize the atom-map to annotate the correspondence between atoms of product and reactants in reaction, which remains unchanged during the reaction. Furthermore, the correspondence of atoms can be readily transferred to the token correspondence in the SMILES sequence (see in Figure~\ref{fig1}c) and construct SMILES Alignment Map (SAM) as:
\begin{equation}
SAM_{i j}= \begin{cases}1 & \text { if } R_{i} \underset{\text { mapped }}{\longleftrightarrow} P_{j} \\ 0 & \text { else }\end{cases}
\end{equation}
Inspired by \citep{guiding}, we similarly employed the guided attention approach by introducing guidance loss $\mathcal{L}_{SA}$ to encourage alignment between the cross attention scores in the final layer of the decoder and SMILES Alignment Map, strengthening the attention weights between corresponding tokens.

\begin{table*}[t]
\caption{Top-k accuracy for retrosynthesis prediction on USPTO-50K. Results for all baseline methods are taken from the respective original publications; ConfRetro results are from this work. Best performance is in \textbf{bold}.}\label{tab1}
\tabcolsep=0pt
\begin{tabular*}{\textwidth}{@{\extracolsep{\fill}}l@{\extracolsep{\fill}}c@{\extracolsep{\fill}}c@{\extracolsep{\fill}}c@{\extracolsep{\fill}}c@{\extracolsep{\fill}}c@{\extracolsep{\fill}}c@{\extracolsep{\fill}}c@{\extracolsep{\fill}}c@{\extracolsep{\fill}}c@{\extracolsep{\fill}}}
\hline
\multicolumn{1}{@{}l}{\multirow{3}{*}{\textbf{Model}}} & \multicolumn{9}{@{}c@{}}{\textbf{Top-k Accuracy (\%)}} \\ \cline{2-10} %
\multicolumn{1}{@{}l}{} & \multicolumn{4}{@{}c@{}}{\textbf{Reaction class known}} & \textbf{} & \multicolumn{4}{@{}c@{}}{\textbf{Reaction class unknown}} \\ \cline{2-5} \cline{7-10} 
\multicolumn{1}{@{}l}{} & 1 & 3 & 5 & 10 & & 1 & 3 & 5 & 10 \\ \hline 
\textbf{Template-Based} & & & & & & & & & \\ \hline 
RetroSim \citep{retrosim} & 52.9 & 73.8 & 81.2 & 88.1 & & 37.3 & 54.7 & 63.3 & 74.1 \\
GLN \citep{gln} & \textbf{64.2} & 79.1 & 85.2 & 90.0 & & 52.5 & 69.0 & 75.6 & 83.7 \\
LocalRetro \citep{localretro} & 63.9 & \textbf{86.8} & \textbf{92.4} & \textbf{96.3} & & \textbf{53.4} & \textbf{77.5} & \textbf{85.9} & \textbf{92.4} \\ \hline 
\textbf{Semi-Template-Based} & & & & & & & & & \\ \hline 
RetroXpert \citep{xpert} & 62.1 & 75.8 & 78.5 & 80.9 & & 50.4 & 61.1 & 62.3 & 63.4 \\
G2G \citep{g2gs} & 61.0 & 81.3 & 86.0 & 88.7 & & 48.9 & 67.6 & 72.5 & 75.5 \\
GraphRetro \citep{graphretro} & 63.9 & 81.5 & 85.2 & 88.1 & & 53.7 & 68.3 & 72.2 & 75.5 \\
RetroPrime \citep{retroprime} & 64.8 & 81.6 & 85.0 & 86.9 & & 51.4 & 70.8 & 74.0 & 76.1 \\
MEGAN \citep{megan} & 60.7 & 82.0 & 87.5 & 91.6 & & 48.1 & 70.7 & 78.4 & 86.1 \\
MARS \citep{liu2024mars} &66.2 & 85.8 & 90.2 & 92.9 & & 54.6 & 76.4 & 83.3 & 88.5 \\
Graph2Edits \citep{graph2edit} & \textbf{67.1} & \textbf{87.5} & \textbf{91.5} & \textbf{93.6} & & \textbf{55.1} & \textbf{77.3} & \textbf{83.4} & \textbf{89.4} \\ \hline 
\textbf{Template-Free} & & & & & & & & & \\ \hline 
SCROP \citep{scrop} & 59.0 & 74.8 & 78.1 & 81.1 & & 43.7 & 60.0 & 65.2 & 68.7 \\
Aug. Transformer \citep{augtrans} & - & - & - & - & & 48.3 & - & 73.4 & 77.4 \\
GTA \citep{gta} & - & - & - & - & & 51.1 & 67.6 & 74.8 & 81.6 \\
Graph2SMILES \citep{graph2smiles} & - & - & - & - & & 52.9 & 66.5 & 70.0 & 72.9 \\
Retroformer \citep{retroformer} & 64.0 & 82.5 & 86.7 & 90.2 & & 53.2 & 71.1 & 76.6 & 82.1 \\
NAG2G \citep{nag2g} & 67.2 & 86.4 & 90.5 & 93.8 & & 55.1 & 76.9 & 83.4 & 89.9 \\ 
\textbf{ConfRetro (Ours)} & \textbf{68.2} & \textbf{88.9} & \textbf{92.8} & \textbf{96.1} & & \textbf{56.2$\pm$0.2}\textsuperscript{\dag} & \textbf{79.3$\pm$0.4} & \textbf{86.2$\pm$0.4} & \textbf{91.7$\pm$0.2} \\
\hline 
\end{tabular*}
\textsuperscript{\dag} Bootstrap 95\% CI for Top-1 (reaction class unknown): [54.8, 57.5] (percentile bootstrap, $B$=10{,}000 over test reactions; see Supplementary Material~\ref{sec:sig} for full results).
\end{table*}

\subsection{Overall Loss}
The overall loss is composed of the language modeling R-Drop loss~\citep{rdrop} $\mathcal{L}_{LM}$ for retrosynthesis, along with the SMILES Alignment guidance loss $\mathcal{L}_{SA}$. 

The objectives of overall loss: i) approximate target in retrosynthesis, ii) ensure the robustness under different dropout, iii) strengthen the alignment relationship in cross-attention. The formula is as follows:
\begin{equation}
\mathcal{L} = \mathcal{L}_{LM}^{(C E)} + \alpha \mathcal{L}_{LM}^{(K L)} + \beta \mathcal{L}_{SA},
\end{equation}
where $\mathcal{L}_{LM}^{(C E)}$ is Cross-Entropy loss to minimize prediction error against actual outputs, $\mathcal{L}_{LM}^{(K L)}$ is KL-divergence in R-Drop loss to strive for outputs of models with varying Dropout to be as similar as possible. $\mathcal{L}_{SA}$ is Cross-Entropy loss between cross-attention score of last decoder layer and SAM that we mention at Section~\ref{sec4.3}. We set $\alpha$ to 0.5, $\beta$ to 1.0.

\section{Experiments}
\subsection{Datasets.} 
We evaluate our model on the widely used retrosynthesis benchmark dataset USPTO-50K\footnote{USPTO-50K and USPTO-FULL datasets can be obtained at \underline{https://github.com/Hanjun-Dai/GLN}}
 \citep{uspto50k}, which contains 50,016 atom-mapped reactions. Following \citep{retrosim}, the data are split into 40,008, 5,001, and 5,007 reactions for the training, validation, and test sets, respectively. Molecular conformers are generated using {\tt RDKit} as follows: each SMILES is parsed with {\tt MolFromSmiles}, hydrogen atoms are added via {\tt AddHs}, a single 3D conformer is embedded with {\tt AllChem.EmbedMolecule} (random seed fixed at 10; up to 10 retries on failure), followed by MMFF94 force-field optimisation with {\tt AllChem.MMFFOptimizeMolecule} (RDKit default, \texttt{maxIters}=200), and finally hydrogen atoms are removed with {\tt RemoveHs}. The same deterministic conformer (seed=10) is used consistently at both training and test time. Stereochemistry specified in the input SMILES ({\tt @}/{\tt @@}, E/Z) is enforced during conformer generation via {\tt MolFromSmiles}; for atoms with unspecified stereochemistry, RDKit assigns an arbitrary but deterministic 3D configuration under the fixed random seed. The robustness of performance to this single-conformer choice is examined in Supplementary Material~\ref{sec:confseed}, and a stereochemistry-stratified evaluation is reported in Supplementary Material~\ref{sec:stereo}. The ground-truth SMILES Alignment Map is extracted using the algorithm described in \citep{retroformer}.
In addition, we conduct experiments on the larger USPTO-FULL dataset\citep{gln}, which comprises 1,013,118 atom-mapped reactions without reaction-class information. This more extensive collection not only enables a comprehensive evaluation of the proposed method’s performance, generalization potential, and scalability but also provides samples with more intricate molecular structures, allowing further validation of our model’s capability to handle complex spatial configurations.

\subsection{Setting.}
Our model is built on vanilla Transformer, consisting of 6 encoder layers and 6 decoder layers with 8 attention heads. The model dimension is set to 512. We employ the Adam optimizer \citep{adam} with $(\beta_1, \beta_2)=(0.9, 0.98)$ and train the model with batch size of 16. During training, we set an early-stop of 7 epochs under a maximum of 1000 epochs to obtain the 7 best models and average their parameters. The choice of averaging 7 checkpoints is empirical as shown in the Supplementary Material~\ref{sec:ckpt}. Finally, the whole training process took approximately 20 hours on a single NVIDIA GeForce RTX 4090 GPU for USPTO-50K; training on USPTO-FULL (approximately 20$\times$ larger) takes approximately one week on the same hardware. More details can be found in Supplementary Material~\ref{sec:appe}.

\subsection{Evaluation.}
During inference, we use beam search \citep{beamsearch} decoding strategy with a beam size of 10. After that, we evaluate our model with \textbf{Top-k Accuracy}, which considers it correct only if Top-k predicted reactants of beam search results are the same as those in the original test set. We verify \textbf{Top-k Validity} based on whether {\tt RDKit} can recognize the Top-k predicted reactants. \textbf{Top-k Round-Trip Accuracy} measures the percentage of predicted reactants that can lead back to the original product.

\subsection{Baseline.}
We compare our proposed method, ConfRetro, with several strong baseline models from three types of retrosynthesis methods. We take RetroSim \citep{retrosim}, GLN \citep{gln} and LocalRetro \citep{localretro} to represent template-based methods. We take RetroXpert \citep{xpert}, G2Gs \citep{g2gs}, GraphRetro \citep{graphretro}, RetroPrime \citep{retroprime}, MEGAN \citep{megan}, MARS\citep{liu2024mars} and Graph2Edits \citep{graph2edit} to represent semi-template methods. We take SCROP \citep{scrop}, Aug.Transformer \citep{augtrans}, Graph2SMILES \citep{graph2smiles}, GTA \citep{gta}, Retroformer \citep{retroformer} and NAG2G \citep{nag2g} to represent template-free methods. We did not choose pre-trained methods for comparison as they are trained on larger datasets. We apply Root-align data augmentation \citep{rsmiles} and conduct experiments on the proposed model
ConfRetro, choosing on-the-fly augmentation instead of directly
expanding the training dataset; following the data augmentation
strategy of previous works, we additionally apply 20× augmentation
on the test set of USPTO-50K and 5× augmentation on the test set
of USPTO-FULL.

\subsection{Performance}
\subsubsection{Performance On USPTO-50K}
\paragraph{\textbf{Top-k Accuracy.}}
As shown in Table~\ref{tab1}, when the reaction class is known, ConfRetro achieves Top1-10 results that outperform all template-free/semi-template baselines and most state-of-the-art template-based baselines. Notably, our model exceeds both template-based and semi-template methods, despite their reliance on predefined templates or RDKit editing. When the reaction class is unknown, our model still achieves improvements in Top1-3 accuracy within the template-free category. Achieving high-confidence predictions (\emph{i.e.}, Top-1 accuracy), which narrows the search space for synthesis routes and enhances efficiency, makes our model particularly well-suited for real-world scenarios. Performance across each reaction class in USPTO-50K dataset is provided in Supplementary Material~\ref{appendixb1}.

\begin{table*}[t]
\caption{Top-k SMILES Validity / Round-Trip Accuracy for retrosynthesis prediction on USPTO-50K with reaction class unknown.}
\label{tab2}
\tabcolsep=0pt
\begin{tabular*}{\textwidth}{@{\extracolsep{\fill}}l@{\extracolsep{\fill}}c@{\extracolsep{\fill}}c@{\extracolsep{\fill}}c@{\extracolsep{\fill}}c@{\extracolsep{\fill}}c@{\extracolsep{\fill}}c@{\extracolsep{\fill}}c@{\extracolsep{\fill}}c@{\extracolsep{\fill}}c@{\extracolsep{\fill}}}
\hline %
\multicolumn{1}{@{}l}{\multirow{2}{*}{\textbf{Model}}} & \multicolumn{4}{@{}c@{}}{\textbf{Top-k Validity (\%)}} & & \multicolumn{4}{@{}c@{}}{\textbf{Top-k Round-Trip (\%)}} \\ \cline{2-5} \cline{7-10} %
\multicolumn{1}{@{}l}{} & 1 & 3 & 5 & 10 & & 1 & 3 & 5 & 10 \\ \hline %
Graph2SMILES & 99.4 & 90.9 & 84.9 & 74.9 & & 76.7 & 56.0 & 46.4 & 34.9 \\
RetroPrime & 98.9 & 98.2 & 97.1 & 92.5 & & 79.6 & 59.6 & 50.3 & 40.4 \\
Retroformer & 99.2 & 98.5 & 97.4 & 96.7 & & 78.9 & 72.0 & 67.1 & 57.2 \\
NAG2G & 99.7 & 98.6 & 97.1 & 92.9 & & - & - & - & - \\ 
\textbf{ConfRetro (Ours)} & \textbf{99.8} & \textbf{99.6} & \textbf{99.1} & \textbf{97.0} & & \textbf{90.7} & \textbf{81.3} & \textbf{75.7} & \textbf{69.0} \\
\hline 
\end{tabular*}
\end{table*}

\paragraph{\textbf{Top-k Validity.}}
We take four typical template-free methods, Graph2SMILES, RetroPrime, Retroformer and NAG2G, as baseline models to verify the predicted SMILES validity of our model. We refrained from selecting template-based and semi-template methods, as SMILES constructed based on template information are guaranteed to be valid. In contrast, template-free methods often struggle to generate valid SMILES without extra chemical guidance. As shown in Table~\ref{tab2}, our model is more capable of generating valid SMILES than the baseline models, especially for Top-5 and Top-10 reactants, where it is far ahead of the others. This indicates that incorporating 3D conformer information can assist the model in grasping the structures involved in chemical reactions, thus preventing the generation of SMILES that violate chemical principles.
\paragraph{\textbf{Top-k Round-Trip Accuracy.}}
To assess the accuracy of our predicted synthesis plans, we use the Molecule Transformer \citep{moltrans} as the benchmark reaction prediction model and calculate the Top-k Round-Trip Accuracy. The results in Table~\ref{tab2} clearly show that our model significantly outperforms other baselines, achieving a remarkable 12\% improvement in Top-1 Round-Trip Accuracy compared to the best baseline. A higher Round-Trip Accuracy indicates that the predicted synthesis plans are more chemically valid and executable, as they can be successfully converted back to the target molecules by a forward reaction prediction model.

\begin{table}[t]
\centering
\caption{Top-k accuracy for retrosynthesis prediction on USPTO-FULL without reaction class. Results for all baseline methods are taken from the respective original publications. Best performance is in \textbf{bold}.}
\label{tab1_1}
\begin{tabular}{lcccc}
\toprule
\multicolumn{1}{c}{\multirow{2}{*}{\textbf{Model}}} & \multicolumn{4}{c}{\textbf{Top-k Accuracy (\%)}}                  \\  
\multicolumn{1}{c}{}                                & 1             & 3             & 5             & 10                \\  \midrule
\textbf{Template-Based}                             &               &               &               &                   \\  \midrule
RetroSim                                            & 32.8          & -             & -             & 56.1              \\
GLN                                                 & 35.8          & -             & -             & 60.8              \\
LocalRetro                                          & 39.3          & -             & -             & 63.7              \\  \midrule
\textbf{Semi-Template-Based}                        &               &               &               &                   \\  \midrule
RetroPrime                                          & 44.1          & 59.1          & 62.8          & 68.5              \\ 
MEGAN                                               & 33.6          & -             & -             & 63.9              \\
Graph2Edits                                         & 44.0          & 60.9          & 66.8          & 72.5              \\  \midrule 
\textbf{Template-Free}                              &               &               &               &                   \\  \midrule
Aug. Transformer                                    & 46.2          & -             & -             & 73.3              \\
GTA                                                 & 46.6          & -             & -             & 70.4              \\
Graph2SMILES                                        & 45.7          & -             & -             & 63.4              \\
NAG2G                                               & 49.7          & 64.6          & 69.3          & 74.0              \\ 
\textbf{ConfRetro (Ours)}                           & \textbf{51.7} & \textbf{69.2} & \textbf{74.5} & \textbf{79.7}     \\
\hline
\end{tabular}
\end{table}

\subsubsection{Performance On USPTO-FULL}
To further assess the scalability and generalization of our method in large-scale and structurally diverse chemical space, we evaluate ConfRetro on USPTO-FULL dataset, which contains over one million atom-mapped reactions and does not provide reaction-class labels. Beyond its scale, USPTO-FULL is typically noisier than USPTO-50K, posing a more challenging setting for robust retrosynthesis modeling. Compared with USPTO-50K, USPTO-FULL exhibits substantially greater reaction diversity and more complex scaffolds, including bulky ring systems, dense stereocenters, and coupled functional-group transformations, making it a stricter stress test for retrosynthesis models.

Table~\ref{tab1_1} reports the Top-$k$ accuracy of ConfRetro and representative baselines spanning template-based, semi-template, and template-free families. Overall, ConfRetro achieves state-of-the-art performance within the template-free category, reaching 51.7\%, 69.2\%, 74.5\%, and 79.7\% Top-1/3/5/10 accuracy, respectively. Notably, it surpasses NAG2G, the strongest prior template-free system, by +2.0\%, +4.6\%, +5.2\%, and +4.3\% at Top-1/3/5/10, respectively. These consistent gains across $k$ suggest that ConfRetro not only improves the confidence of its top-ranked prediction but also yields a higher-quality candidate set under beam search.

It is also worth emphasizing that ConfRetro remains competitive even when compared with template-based and semi-template approaches such as LocalRetro~\citep{localretro} and Graph2Edits~\citep{graph2edit}. Unlike these methods, ConfRetro is trained end-to-end without predefined reaction templates or hard-coded edit operations. By integrating 3D conformer information and explicit product-reactant alignment supervision, the model better accounts for spatial constraints, improving robustness on sterically congested and geometrically intricate motifs that are often ambiguous from 2D topology alone.

We attribute the performance gains on USPTO-FULL to two main factors. First, conformer-aware encodings (Atom-align Fusion and Distance-weighted Attention) increase sensitivity to spatial proximity and steric coupling, which becomes increasingly important for large-ring and polyfunctional molecules. Second, SMILES Alignment supervision stabilizes decoding for long and heterogeneous products by encouraging consistent structural correspondence between product and reactant tokens, mitigating beam-search degeneration and improving the quality of alternatives at larger $k$.

Overall, the USPTO-FULL results demonstrate that ConfRetro scales beyond curated benchmarks and remains reliable in a setting that more closely reflects real-world retrosynthesis workloads, characterized by the absence of reaction-class hints, high transformation diversity, increased noise, and structurally complex targets.

\begin{table*}[t]
\caption{Ablation study of ConfRetro with reaction class unknown on USPTO-50K dataset. \textbf{Module (a)} is Atom-align Fusion; \textbf{Module (b)} is Distance-weighted Attention and \textbf{Module (c)} is SMILES Alignment;}
\label{tab3}
\tabcolsep=0pt
\begin{tabular*}{\textwidth}{@{\extracolsep{\fill}}c@{\extracolsep{\fill}}c@{\extracolsep{\fill}}c@{\extracolsep{\fill}}c@{\extracolsep{\fill}}c@{\extracolsep{\fill}}c@{\extracolsep{\fill}}c@{\extracolsep{\fill}}c@{\extracolsep{\fill}}c@{\extracolsep{\fill}}c@{\extracolsep{\fill}}c@{\extracolsep{\fill}}c@{\extracolsep{\fill}}c@{\extracolsep{\fill}}}
\hline 
\multicolumn{1}{@{}c@{}}{} & \multicolumn{1}{@{}c@{}}{\multirow{2}{*}{\textbf{(a)}}} & \multicolumn{1}{@{}c@{}}{\multirow{2}{*}{\textbf{(b)}}} & \multicolumn{1}{@{}c@{}}{\multirow{2}{*}{\textbf{(c)}}} & \multicolumn{4}{@{}c@{}}{\textbf{Top-k Accuracy (\%)}} & & \multicolumn{4}{@{}c@{}}{\textbf{Top-k Validity (\%)}} \\ \cline{5-8} \cline{10-13} 
\multicolumn{1}{@{}c@{}}{} & \multicolumn{1}{@{}c@{}}{} & \multicolumn{1}{@{}c@{}}{} & \multicolumn{1}{@{}c@{}}{} & 1 & 3 & 5 & 10 & & 1 & 3 & 5 & 10 \\ \hline 
\ding{172} & - & - & - & 50.9 & 72.7 & 78.8 & 82.7 & & 99.1 & 98.4 & 97.7 & 95.6 \\
\ding{173} & - & - & $\checkmark$ & 53.5 & 75.9 & 83.5 & 89.0 & & 99.3 & 98.9 & 98.5 & 95.3 \\
\ding{174} & $\checkmark$ & - & $\checkmark$ & 54.0 & 75.2 & 81.9 & 87.2 & & 99.6 & 99.2 & 98.4 & 94.9 \\
\ding{175} & - & $\checkmark$ & $\checkmark$ & 54.8 & 75.3 & 81.9 & 87.8 & & 99.7 & 99.4 & 98.6 & 94.2 \\
\ding{176} & $\checkmark$ & $\checkmark$ & - & 54.8 & 75.1 & 82.8 & 89.1 & & 99.7 & 99.6 & 98.9 & 95.9 \\ 
\ding{177} & $\checkmark$ & $\checkmark$ & $\checkmark$ & \textbf{56.2} & \textbf{79.3} & \textbf{86.2} & \textbf{91.7} & & \textbf{99.8} & \textbf{99.6} & \textbf{99.1} & \textbf{97.0} \\
\hline 
\end{tabular*}
\end{table*}

\subsection{Ablation Study}
To quantify the contribution of each proposed component, we perform an ablation study on USPTO-50K under the reaction-class-unknown setting. Table~\ref{tab3} reports the Top-$k$ accuracy and SMILES validity of different model variants denoted by \ding{172} to \ding{177}, corresponding to different combinations of modules (a) Atom-align Fusion, (b) Distance-weighted Attention, and (c) SMILES Alignment.

Comparing \ding{173} with the baseline \ding{172}, introducing SMILES Alignment (c) leads to a consistent improvement in Top-$k$ accuracy, indicating that explicit alignment supervision stabilizes decoding under beam search. Further incorporating Atom-align Fusion (a) (\ding{174} vs. \ding{173}) improves Top-1 accuracy, suggesting that conformer-informed atom-level correspondence helps the model better prioritize the most plausible reactant set. In contrast, adding Distance-weighted Attention (b) (\ding{175} vs. \ding{173}) yields gains across all Top-$k$ values, with more pronounced improvements at larger $k$, reflecting enhanced modeling of internal molecular structure.

The full model \ding{177}, which combines all three modules, achieves the best overall performance, simultaneously improving high-confidence predictions and maintaining strong results at larger beam widths. These results demonstrate that conformer-aware encoding and alignment-based supervision are complementary and jointly contribute to the effectiveness of ConfRetro.

\begin{table}[t]
\caption{Sources of improvement on USPTO-50K (reaction class unknown). Each row varies a single factor relative to ConfRetro (ours): w/o bond-type fixes the bond-type affine to identity, w/o 3D distance zeroes the distance term, w/ random conformer replaces coordinates with noise, and w/o TTA disables test-time augmentation.}
\label{tab_sources}
\resizebox{\columnwidth}{!}{%
\begin{tabular}{lcccc}
\toprule
\textbf{Variant} & \textbf{Top-1} & \textbf{Top-3} & \textbf{Top-5} & \textbf{Top-10} \\
\midrule
w/o bond-type & 56.0 & 79.2 & \textbf{86.5} & \textbf{91.8} \\
w/o 3D distance & 53.0 & 77.4 & 84.5 & 90.5 \\
w/ random conformer & 53.4 & 77.7 & 84.8 & 90.7 \\
w/o TTA ($n_{\text{aug}}$=1) & 55.5 & 77.2 & 83.4 & 89.1 \\
\midrule
\textbf{ConfRetro (ours)} & \textbf{56.2} & \textbf{79.3} & 86.2 & 91.7 \\
\bottomrule
\end{tabular}%
}
\end{table}

To further investigate \emph{why} the conformer-aware modules help, we disentangle 3D geometry from model capacity, atom-level supervision, bond-type features, and test-time augmentation by varying a single factor at a time while holding all other components constant. Table~\ref{tab_sources} reports the results on USPTO-50K (reaction class unknown).

Three findings emerge. \emph{(i) 3D geometry is the causal source.} Replacing the real conformer with random coordinates drops Top-1 accuracy by 2.7 points (56.2 $\rightarrow$ 53.4). Since every factor other than the atomic coordinates is held fixed, this gap is attributable to genuine 3D geometry rather than added parameters or supervision, and is statistically significant (McNemar $p$=$9.0\!\times\!10^{-8}$). \emph{(ii) The benefit comes from distance, not bond type.} The distance module combines an inter-atomic distance term with a learnable bond-type affine. Removing the bond-type affine (w/o bond-type) leaves performance essentially unchanged (56.0 vs.\ 56.2 Top-1), whereas removing the distance term (w/o 3D distance) collapses accuracy to 53.0, comparable to the random-conformer control, so the useful signal lies in the actual 3D distances. \emph{(iii) The gains are not an artefact of augmentation.} Even with test-time augmentation disabled ($n_{\text{aug}}$=1), ConfRetro still attains Top-1 accuracy of 55.5\%, well above both controls. Bootstrap confidence intervals and paired significance tests for all variants are reported in Supplementary Material~\ref{sec:sig}.

\begin{figure*}
  \centering
  \subfigure[\footnotesize Polychiral]{
    \includegraphics[width=0.47\linewidth]{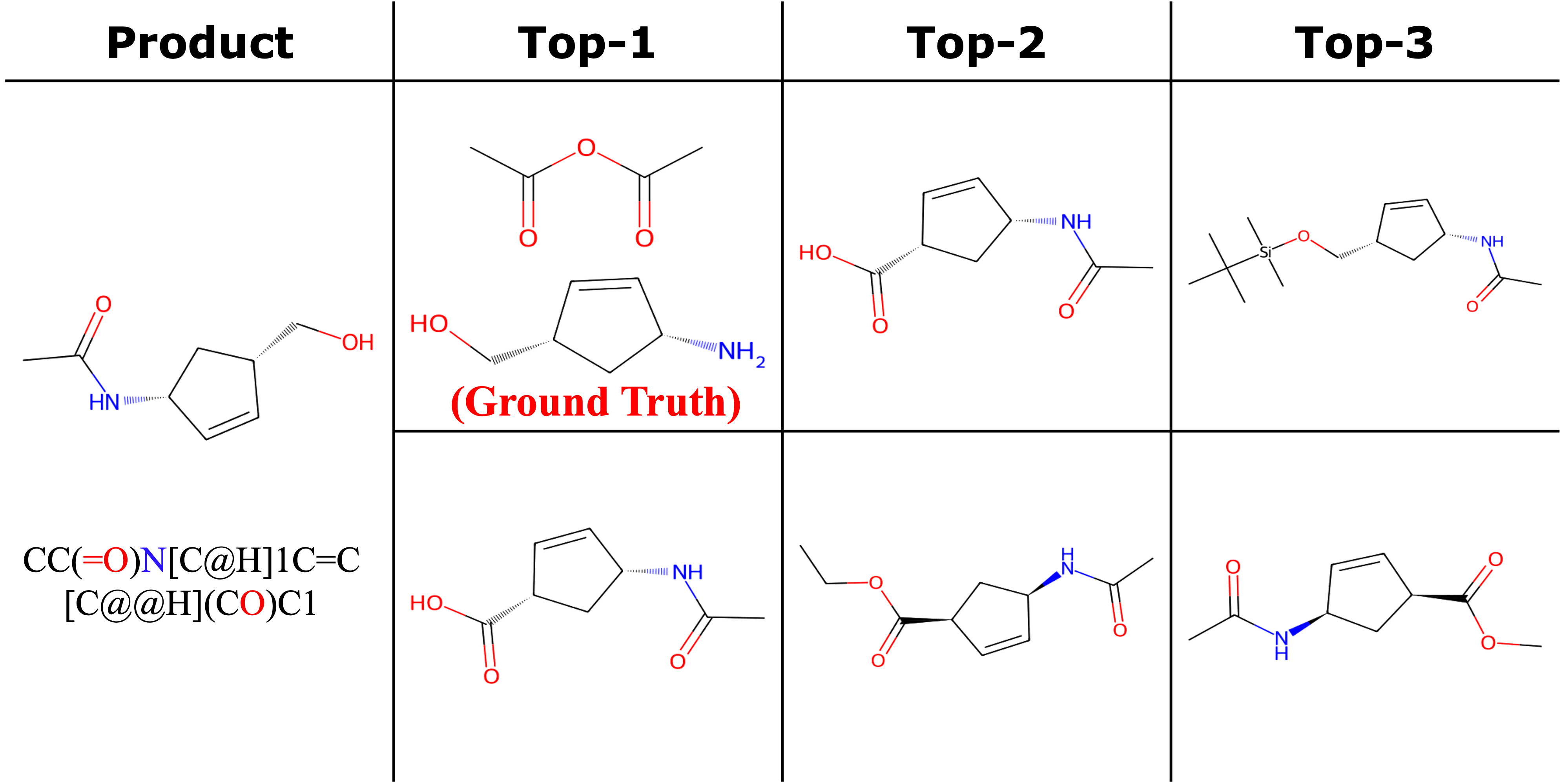}
    \label{fig2a}
  }
  \hfill
  \subfigure[\footnotesize Heteroaromatic]{
    \includegraphics[width=0.47\linewidth]{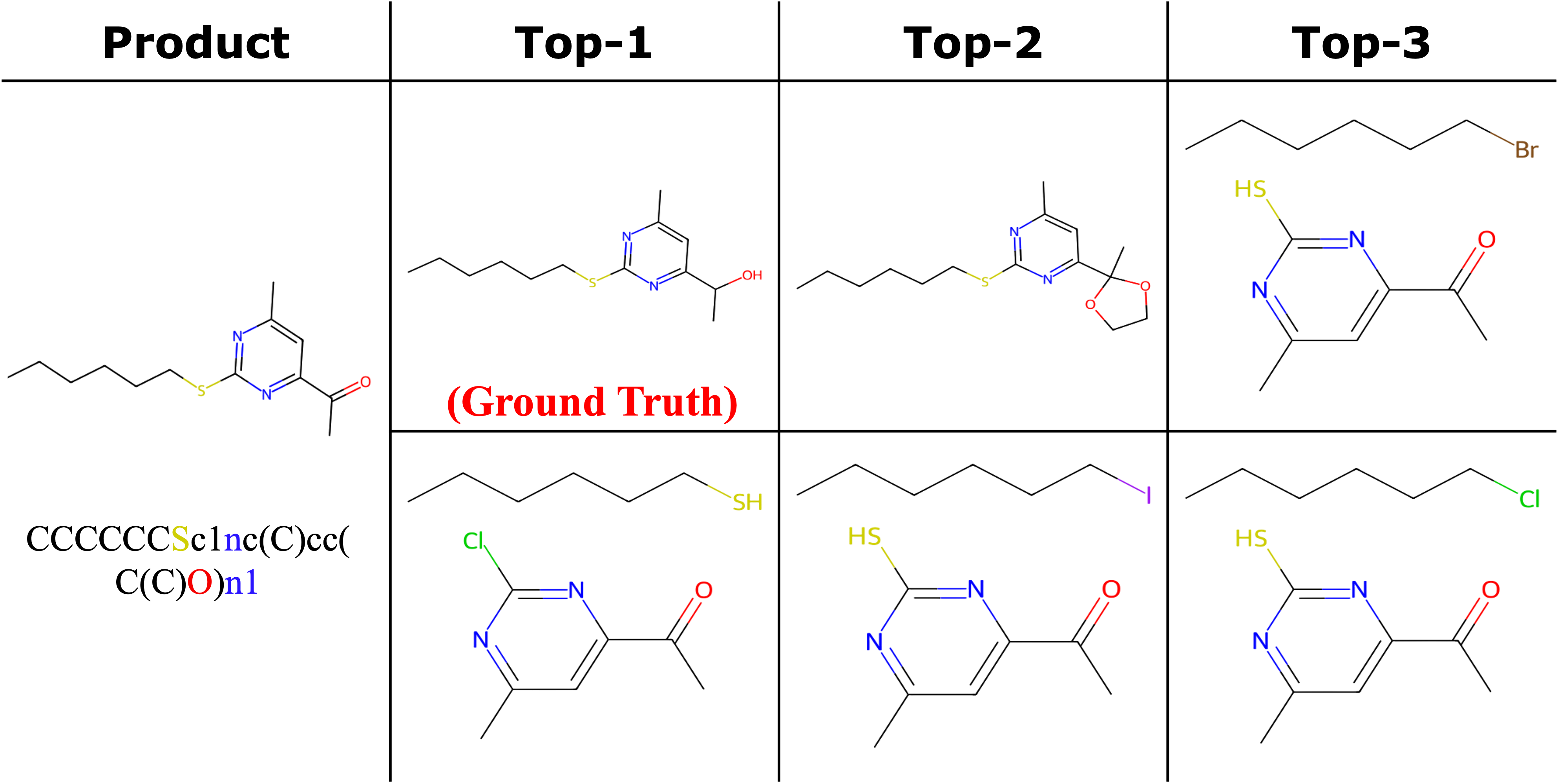}
    \label{fig2b}
  }

  \subfigure[\footnotesize Fused / Bridged Rings]{
    \includegraphics[width=0.47\linewidth]{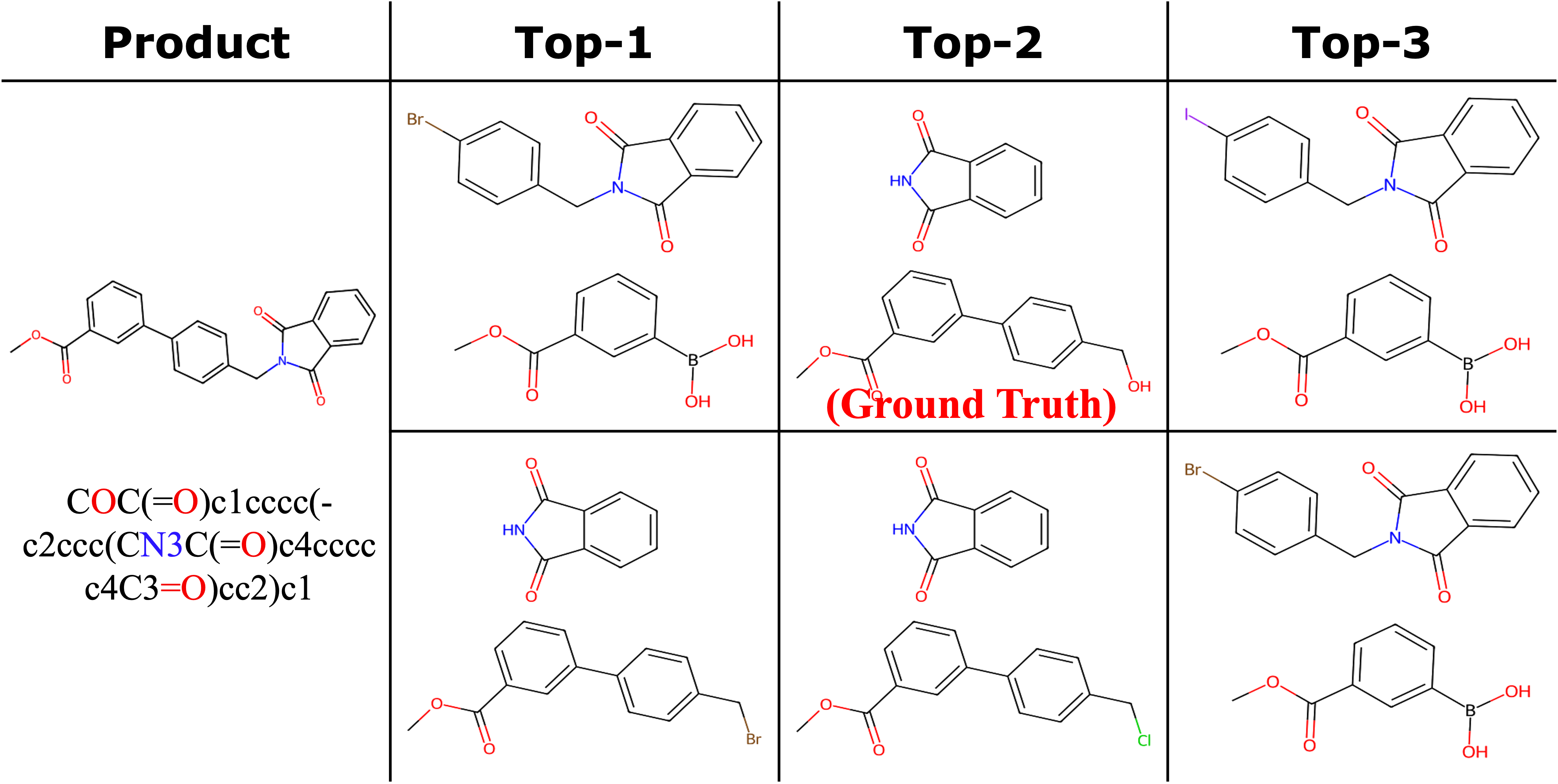}
    \label{fig2c}
  }
  \hfill
  \subfigure[\footnotesize Complex]{
    \includegraphics[width=0.47\linewidth]{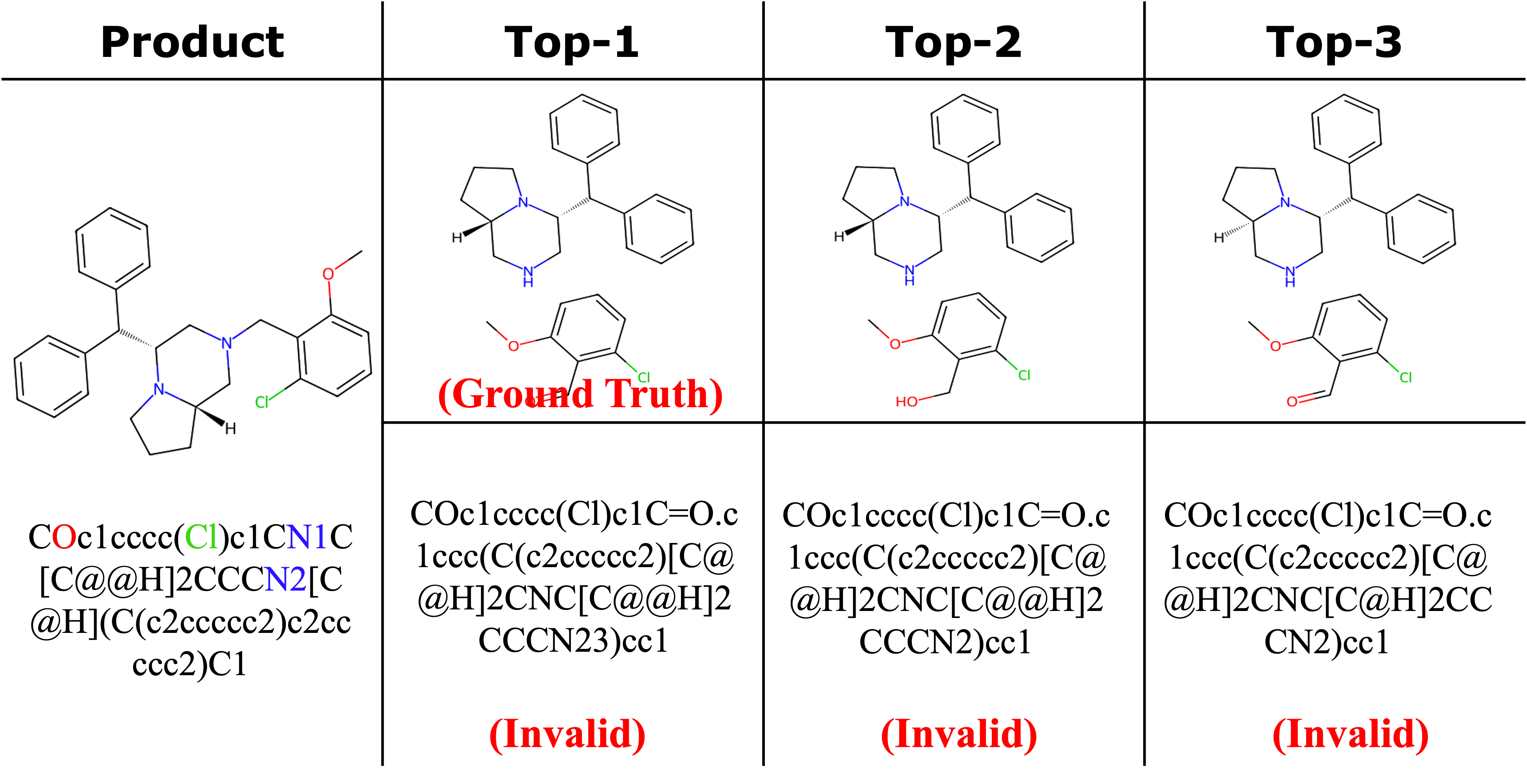}
    \label{fig2d}
  }

  \caption{Case study for intricate structures. Each subfigure shows predictions of different methods: \textbf{ConfRetro (Top)}, \textbf{Transformer (Bottom)}.}
  \label{fig2}
\end{figure*}

\subsection{Case Study}
To verify that the proposed method ConfRetro can handle molecules with more intricate spatial structures, we chose 4 types of products as representatives:

\begin{itemize}
    \item Polychiral: a product with two or more chiral centers ('{\tt @/@@}' in SMILES), leading to complex stereochemistry and multiple isomers.
    \item Heteroaromatic: a product containing at least one heteroatom as part of its aromatic ring structure.
    \item Fused / Bridged Rings: product features multiple ring structures joined together, either sharing common atoms or connected by bridge atoms.
    \item Complex: a product encompassing two or more features mentioned above.
\end{itemize}

\begin{figure}[t]
    \centering
    \includegraphics[width=0.9\linewidth]{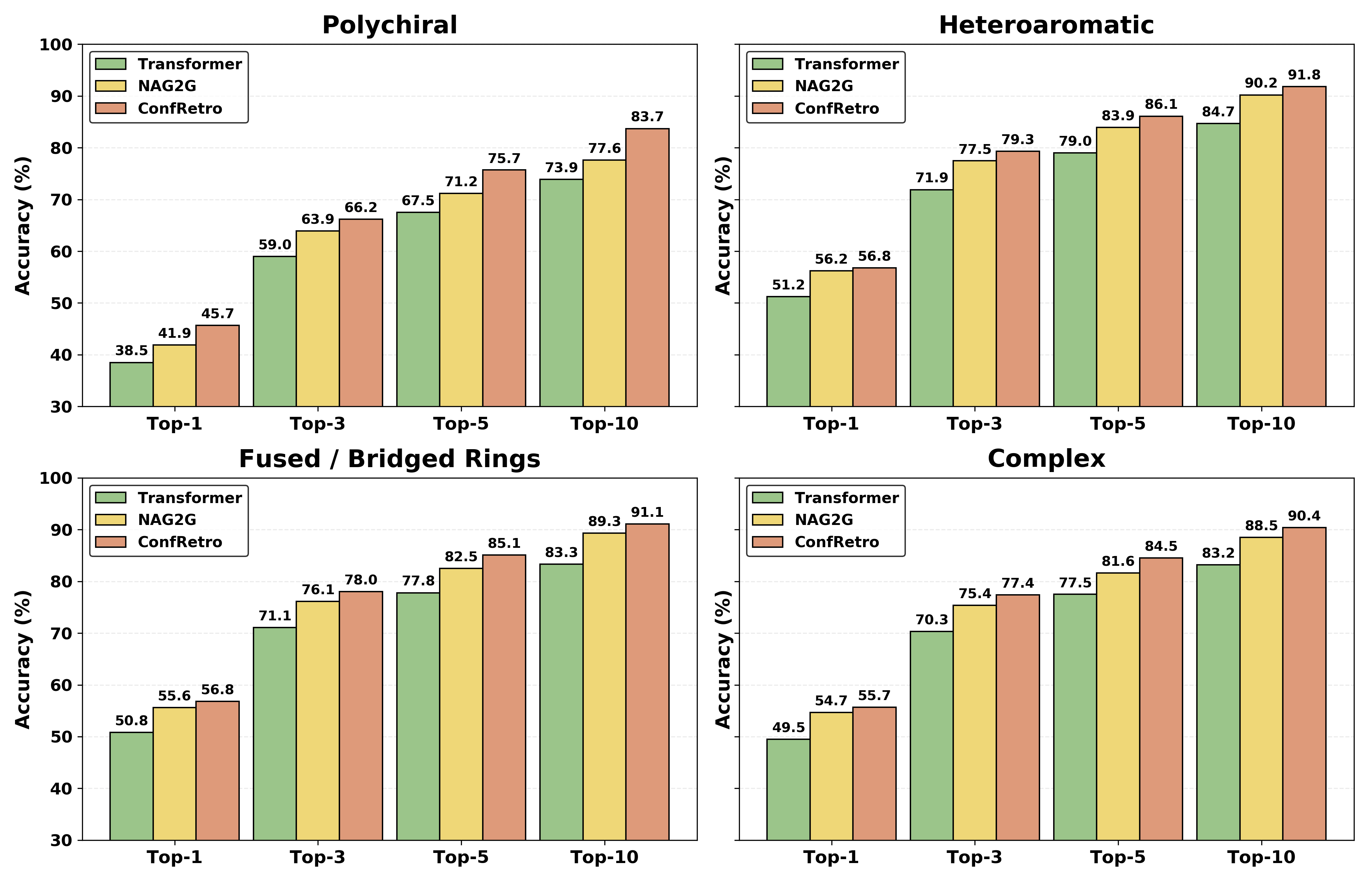}
    \caption{Quantitative comparison of ConfRetro, NAG2G, and Vanilla Transformer on four types of complex molecules.}
    \label{fig3}
\end{figure}

We conducted experiments comparing ConfRetro with a Vanilla Transformer, analyzing Top1-3 predictions, as illustrated in Figure~\ref{fig2}. The rows in each subfigure represent the results of ConfRetro and Transformer, respectively, with ground truth and invalid SMILES highlighted in bold red.

The results show that when dealing with molecules of intricate spatial structures, ConfRetro can infer reasonable reactants with high confidence (often at Top-1 or Top-2), while the Transformer without 3D information struggles to generate accurate predictions. Notably, for Complex products, ConfRetro successfully deduces the molecular skeleton and achieves the ground truth at Top-1, while the Vanilla Transformer fails to produce any valid SMILES within the Top-3 predictions.

To further demonstrate the effectiveness of ConfRetro, we conducted a quantitative comparison across all four categories against both the Vanilla Transformer and NAG2G, another 3D-aware template-free method. As shown in Figure~\ref{fig3}, ConfRetro consistently outperforms both baselines across all categories and all Top-$k$ values. Importantly, both 3D-aware methods (ConfRetro and NAG2G) outperform the 2D Transformer, yet ConfRetro surpasses NAG2G in all four categories, indicating that the performance gain stems from \emph{how} 3D geometry is used (specifically, the Atom-align Fusion and Distance-weighted Attention design) rather than simply having 3D inputs available. The performance gap is most pronounced for Polychiral molecules (the hardest category), where incorporating 3D spatial information is crucial for resolving stereochemical ambiguities. Detailed per-category Top-$k$ accuracy and SMILES validity for the three methods are reported in Supplementary Material~\ref{sec:case}.

\section{Synthesis route}

To demonstrate the practical applicability and robustness of our framework in planning feasible multi-step synthetic routes for structurally intricate, real-world pharmaceuticals, we conducted a case study on a set of complex target molecules. A key objective of this experiment was to validate that our method can successfully navigate the challenging chemical search space of polycyclic natural products with stringent stereochemical requirements, thereby generating complete and chemically valid retrosynthetic pathways.

For this purpose, we selected Camptothecin as a representative and highly challenging test case. Camptothecin is a clinically significant anticancer natural product that functions as a potent DNA topoisomerase I inhibitor \citep{Li2016Camptothecin}. Its complex structure and intricate biosynthesis make it a challenging molecule for retrosynthesis.

\begin{figure}[t]
    \centering
    \includegraphics[width=0.99\linewidth]{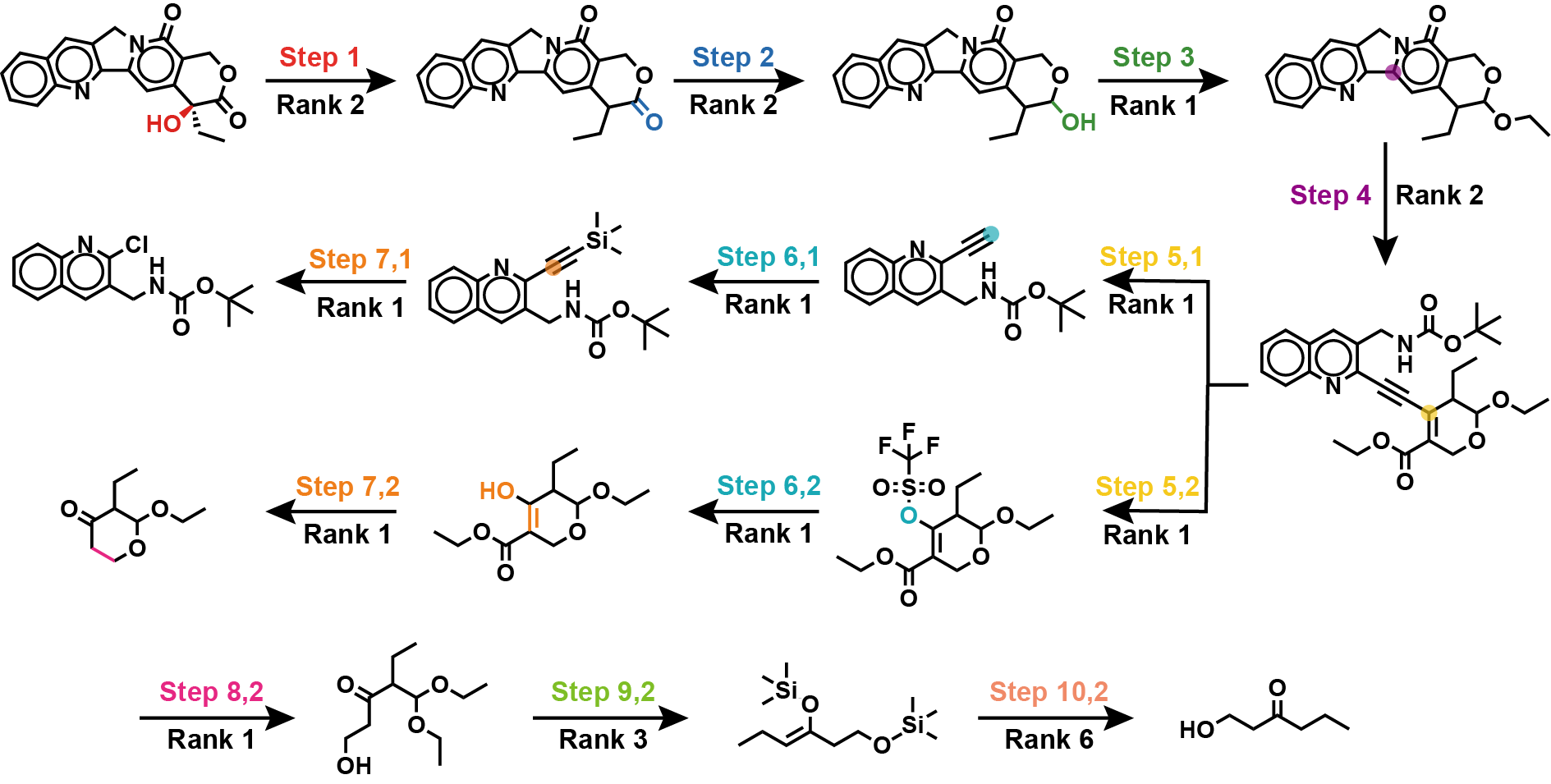}
    \caption{Camptothecin multistep retrosynthesis predictions by our method. The first precursor is 20-deoxycamptothecin; its C20(S) centre is not a chirality error but is set later in the forward synthesis by an asymmetric $\alpha$-hydroxylation \citep{Tagami2000Camptothecin}.}
    \label{fig4}
\end{figure}

As shown in Figure~\ref{fig4}, ConfRetro was applied iteratively as a single-step predictor to deconstruct camptothecin. The proposed pathway aligns closely with the literature synthesis \citep{Li2016Camptothecin}, confirming its feasibility. This robustness stems from our conformer-aware design, particularly Atom-align Fusion and Distance-weighted Attention, which enforce 3D structural and steric constraints during decoding.

\begin{figure}[t]
    \centering
    \includegraphics[width=0.99\linewidth]{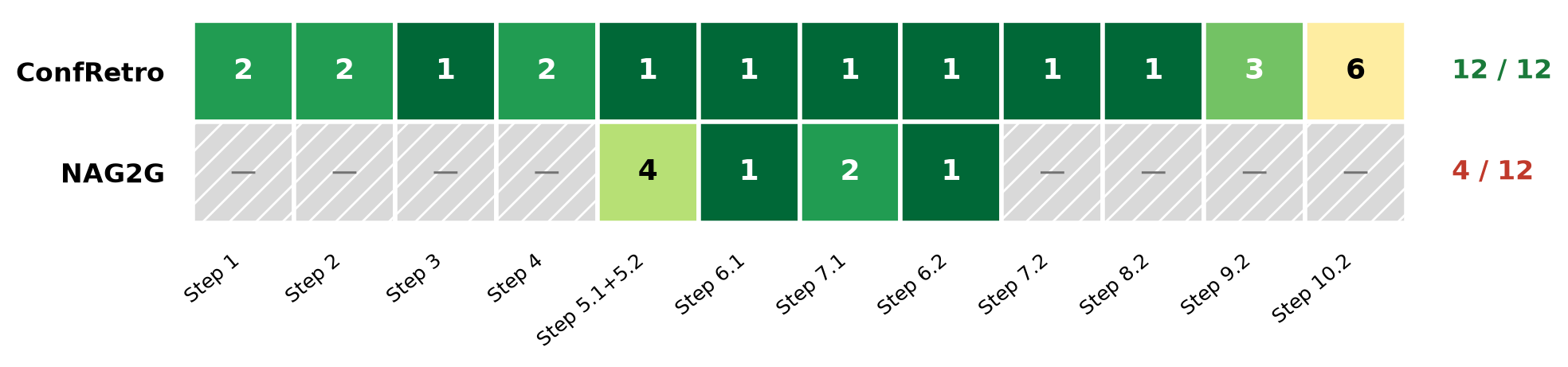}
    \caption{ConfRetro vs.\ NAG2G on the camptothecin route. Each cell gives the rank of the literature precursor at that step (lower is better; a dash denotes not recovered within the top-10 candidates).}
    \label{fig5}
\end{figure}

To further assess whether the performance advantage stems from the proposed 3D design rather than from using 3D inputs per se, we compare ConfRetro against NAG2G, another 3D-aware template-free method, on the same camptothecin route. As shown in Figure~\ref{fig5}, on this particular route ConfRetro recovers all 12 retrosynthetic steps within the top-6 predictions, whereas NAG2G recovers 4 steps, which are mostly the near-terminal building-block disconnections, and tends to miss several of the earlier ring-forming disconnections of the fused polycyclic core. While this is a single illustrative case rather than a quantitative benchmark, the observation is consistent with our single-step results and suggests that ConfRetro's advantage may stem from how 3D information is integrated rather than merely from the availability of 3D inputs.

The results underscore the strength of our approach in handling intricate molecular architectures and its potential utility in accelerating the discovery and development of novel therapeutics. Additional synthesis routes for other representative molecules are provided in Supplementary Material~\ref{sec:synroute}.

\section{Conclusion}
We propose ConfRetro, an end-to-end Transformer that integrates 3D conformer information for retrosynthesis. With the proposed Atom-align Fusion module, we can adaptively integrate token and 3D position information while ensuring the alignment between them. Next, we propose a Distance-weighted Attention mechanism to guide and refine the redistribution of self-attention with spatial distances.
Our model 
achieves a new state-of-the-art in template-free methods and is also highly competitive with template-based and semi-template methods. 
Furthermore, to validate the practicality of the model, we predicted synthesis routes for various candidate drug molecules, demonstrating its potential to revolutionize the prediction of complex chemical synthesis processes in future synthesis route design tasks. The results are provided in
Supplementary Material~\ref{sec:synroute}.

\section*{Competing interests}
No competing interest is declared.

\section*{Author contributions statement}


J.Z. and Y.Z. conceived the study, designed the methodology and conducted the experiments. J.Z. implemented the model and performed the ablation and case studies. Y.Z. contributed to data preparation and experimental evaluation. Y.Q. and A.Z. supervised the project, provided feedback and acquired funding.

\section*{Acknowledgments}
Thanks the anonymous reviewers for their valuable suggestions.
This work is supported by the National Key Research and Development Program of China (2024YFC3308500) and Shanghai Frontiers Science Center of Molecule Intelligent Syntheses.

\bibliographystyle{abbrvnat}
\bibliography{reference}

\onecolumn
\setlength{\parindent}{0pt} 
\setlength{\parskip}{0.5em} 

\setcounter{section}{0}

\renewcommand{\thesection}{\Alph{section}}

\renewcommand{\thesubsection}{\thesection\arabic{subsection}}

\renewcommand{\theHsection}{appendix.\thesection}
\renewcommand{\theHsubsection}{appendix.\thesubsection}

\section{Related Work}

\textbf{Template-based methods} utilize reaction templates from pre-constructed database to capture the rules of chemical changes and match input products with the templates. RetroSim \citep{retrosim} is a similarity-based approach using molecular fingerprint similarity to rank candidate templates. GLN \citep{gln} employs the Conditional Graph Logic Network to learn when rules from reaction templates should be applied, implicitly considering chemical feasibility and strategy. Inspired by the chemical intuition that molecular changes primarily occur locally during reactions, LocalRetro \citep{localretro} is locally encoded and refined to accommodate non-local reactions. Although these methods have strong interpretability, their reliance on pre-defined templates limits their coverage and scalability, rendering them insufficient for real-world scenarios.

\textbf{Semi-template methods} do not directly use templates but instead combine chemical rules with generative models. Existing methods typically follow a two-stage procedure: i) reaction center identification; ii) synthons completion. Reaction center identification predicts the reaction center of the product and breaks the corresponding bond into incomplete molecules, called synthons. Synthons completion uses generative models to complete synthons into reactants. RetroXpert \citep{xpert} builds an Edge-enhanced Graph Attention Network (EGAT) that takes a molecule graph as input and predicts bond disconnection probabilities to obtain synthons. It then employs a seq2seq model to generate reactants SMILES. In contrast to SMILES format, G2Gs \citep{g2gs} completes synthons into reactants in graph format through node selection and edge labeling. GraphRetro \citep{graphretro} expands synthons into complete reactants by attaching leaving groups. RetroPrime \citep{retroprime} introduces a two-stage procedure by adding additional labels to SMILES. In contrast, MEGAN \citep{megan} ,MARS\citep{liu2024mars} and Graph2Edits \citep{graph2edit} reframe the generative process as a series of graph edits executed by {\tt RDKit}.

\textbf{Template-free methods} transform retrosynthesis prediction (product$\rightarrow$reactants) into a machine translation task (source language $\rightarrow$ target language) \citep{machinetranslation}, adapting advanced NLP models like LSTM \citep{lstm} and Transformer \citep{transformer}. Besides translation, SCROP \citep{scrop} builds a self-correction transformer to rectify the syntax errors. Augmented Transformer \citep{augtrans} trains the model with data augmentation via SMILES permutation. To ensure the permutation invariance of SMILES, Graph2SMILES \citep{graph2smiles} substitutes the original sequence encoder with a graph encoder. GTA \citep{gta} masks the self-attention layer using the graph adjacency matrix. Retroformer \citep{retroformer} suggests a local attention head that enables information exchange between the local reactive region and the global reaction context. NAG2G \citep{nag2g} incorporates conformer as additional features but overlooks semantic alignment with SMILES and spatial receptive fields, failing to fully exploit the potential of 3D information.

\section{USPTO-50K Dataset}

\subsection{Performance across Each Reaction Class}

\label{appendixb1}

In Table 3, we present the top-$k$ accuracy ($k$=1, 3, 5, 10) of ConfRetro on the USPTO-50k dataset, where the reaction types are unknown during the training process. The results are categorized by the ground truth reaction types.

Our method seems to predict certain reaction types more accurately than other methods, especially those that are more concerned with 3D spatial structure, such as in Heterocycle formation, where the 3D structure of the molecule directly affects the stereoselectivity and yield of the cyclization reaction. It is challenging for other methods to deduce accurate results without 3D structural information.

\begin{table}[!htb]
\centering
\label{app:table1}
\caption{The top-$k$ accuracy of each reaction class on USPTO-50k dataset, when trained with reaction class unknown.}
\begin{tabular}{lcccccccc}
\toprule
\multicolumn{1}{c}{\multirow{2}{*}{\textbf{Reaction Class}}} & \multicolumn{1}{c}{\multirow{2}{*}{\textbf{Reaction Fraction (\%)}}} & \multicolumn{4}{c}{\textbf{Top-$k$ Accuracy (\%)}} \\ 
                                                             &                                                                      & \textbf{1} & \textbf{3} & \textbf{5} & \textbf{10} \\ \midrule
Heteroatom alkylation and arylation                          & 30.3                                                                & 56.6       & 80.4       & 86.0       & 92.0 \\
Acylation and related processes                              & 23.8                                                                 & 68.5       & 89.9       & 94.3       & 97.7 \\
Deprotections                                                & 16.5                                                                 & 54.5       & 78.5       & 85.9       & 91.9 \\
C-C bond formation                                           & 11.3                                                                 & 41.6       & 63.1       & 73.9       & 83.4 \\
Reductions                                                   & 9.2                                                                  & 56.4       & 78.6       & 85.3       & 90.8 \\
Functional group interconversion                             & 3.7                                                                  & 39.4       & 56.4       & 64.6       & 76.1 \\
Heterocycle formation                                        & 1.8                                                                  & 47.0       & 70.7       & 72.9       & 81.8 \\
Oxidations                                                   & 1.6                                                                  & 61.2       & 82.8       & 91.5       & 95.3 \\
Protections                                                  & 1.3                                                                  & 68.1       & 80.5       & 89.4       & 93.9 \\
Functional group addition                                    & 0.5                                                                  & 80.2       & 85.2       & 89.6       & 98.4 \\
\hline
\end{tabular}
\end{table}

\subsection{Performance on Different Conformer Generator}
To explore the impact of different conformer generators on model performance, we conduct an ablation study. As shown in Table~5, in addition to the default {\tt RDkit} generator, we evaluate the performance of the chemical simulation software {\tt Schrödinger}, which is widely regarded as capable of producing higher-quality conformations. The results show that the overall performance remains comparable.
This observation suggests that our model exhibits strong robustness, as the quality of conformations has minimal impact on the final performance of ConfRetro.

\begin{table}[!htb]
\centering
\label{app:table2}
\caption{The performance with different conformer generator on USPTO-50k with reaction class unknown. All variants are evaluated with $n_{\text{aug}}$=1 for fair comparison. The Default ({\tt RDKit}) model achieves 56.2/79.3/86.2/91.7 with $n_{\text{aug}}$=20 protocol (see main text Table~1).}
\begin{tabular}{lcccc}
\toprule
\multicolumn{1}{c}{\multirow{2}{*}{\textbf{Conformer Generator}}} & \multicolumn{4}{c}{\textbf{Top-k Accuracy (\%)}}                  \\  
\multicolumn{1}{c}{}                                & 1             & 3             & 5             & 10                \\  \midrule
{\tt Schrödinger} \citep{confgen}                                         & 55.2          & 77.1          & \textbf{83.8} & \textbf{90.1}     \\
Default ({\tt RDkit})                                     & \textbf{55.5} & \textbf{77.2} & 83.4          & 89.1              \\
\hline
\end{tabular}

\end{table}

\subsection{Performance on Different 3D Representation}
This section explores the impact of 3D representations on model performance. Table~6 shows that as the 3D representation capacity increases (from top to bottom), the model's performance improves, indicating a positive correlation between them.

\begin{table}[!htb]
\centering
\label{app:table3}
\caption{The performance with different 3D representation networks on USPTO-50k with reaction class unknown. All variants are evaluated with $n_{\text{aug}}$=1 for fair comparison. The Default (ComENet) model achieves 56.2/79.3/86.2/91.7 with $n_{\text{aug}}$=20 protocol (see main text Table~1).}
\begin{tabular}{lcccc}
\toprule
\multicolumn{1}{c}{\multirow{2}{*}{\textbf{3D Representation Network}}} & \multicolumn{4}{c}{\textbf{Top-k Accuracy (\%)}}                  \\  
\multicolumn{1}{c}{}                                & 1             & 3             & 5             & 10                \\  \midrule
SchNet \citep{schnet}                                              & 51.0          & 69.5          & 73.5 & 76.3     \\
SphereNet \citep{spherenet}                                              & 53.1          & 77.1          & 83.1 & 87.8     \\
Default (ComENet)                                   & \textbf{55.5} & \textbf{77.2} & \textbf{83.4}          & \textbf{89.1}              \\
\hline
\end{tabular}

\end{table}

\section{USPTO-FULL Datasets}

We also use a more extensive collection, USPTO-FULL Dataset, which contains 1,013,118 atom-mapped reactions without any reaction class information. By experimenting with it, we can evaluate the proposed method's performance, generalization potential, and scalability, offering valuable insights into its practical application in retrosynthesis. Additionally, the dataset includes samples with \textbf{more intricate structures}, which allows for further validation of our model's capability to handle molecules with intricate spatial configurations.

\subsection{Preprocess}

The USPTO-FULL raw data contains errors and other issues, so we implemented the following steps to clean and refine the dataset:

\begin{itemize}
    \item Exclude reactions with invalid SMILES, including those with empty SMILES strings.
    \item Exclude reactions where the product contains atoms not present in the reactants.
    \item Exclude reactants if none of their atoms are found in the product.
    \item Exclude reactions where multiple atoms share the same atom map number.
    \item Exclude reactions where the product is composed of less than five atoms.
\end{itemize}

The final version shows an approximate 4\% reduction compared to the raw data. The dataset is divided into training, validation, and test sets with sizes of 768K, 96K, and 96K, respectively.

\section{Implementation Details}

\subsection{3D Distance Matrix Construction}
In the Distance-weighted Attention mechanism, a 3D Distance Matrix needs to be constructed first. Algorithm~\ref{algorithm1} is the pseudo code for the construction of the 3D Distance Matrix $\mathbf{D}$ from a molecular canonical-SMILES $S$. 

The algorithm begins by converting the SMILES into a molecular object $M$, followed by generating a conformer for the molecule and obtaining the coordinates of all atoms within it. Subsequently, an empty distance matrix is initialized. For each pair of atom tokens derived from tokenizing the SMILES, if both tokens represent atoms, the algorithm calculates the Euclidean distance between these two atoms and assigns this value to the corresponding position in the distance matrix; if they are not atom tokens, the algorithm sets the corresponding matrix element to 0. Finally, the algorithm returns the constructed 3D distance matrix $\mathbf{D}$.

\begin{algorithm}[!htb]
    \caption{3D Distance Matrix Construction}
    \label{algorithm1}
    \textbf{Input}: molecule canonical SMILES $S$\\
    \textbf{Output}:  3D Distance Matrix $\mathbf{D}$
    \begin{algorithmic}[1]
        \State Let $M=\operatorname{MolFromSmiles}(S)$.
        \State Generate conformer by $\operatorname{MMFFOptimizeMolecule}(M)$.
        \State Get coordinates of atoms $\mathbf{c}=\operatorname{GetPositions}(M)$.
        \State Initialize $\mathbf{D}$ as an empty array.
        \For{$t_i, t_j \in \operatorname{SmiTokenizer}(S)$}
            \If {$t_i$, $t_j$ are atom tokens}
                \State $\mathbf{D_{i,j}}=\operatorname{norm}(\mathbf{c_i}-\mathbf{c_j})$
            \Else
                \State $\mathbf{D_{i,j}}=0$
            \EndIf
        \EndFor
        \State \textbf{return} $\mathbf{D}$
    \end{algorithmic}
\end{algorithm}

\subsection{Root-align SMILES Construction}
Algorithm~\ref{algorithm3} is the pseudo code for the construction of the reactants Root-align SMILES $S_R$({\tt +R}) from a reactants canonical SMILES $S$. 

First, the algorithm extracts the atom-mapped number from the product SMILES $S_P$, then iterates through the SMILES for each reactant, looking for the same atom-mapped number as in the $S_P$. Once a match is found, the algorithm uses the {\tt {\tt RDkit}} toolkit to generate a fixed-root SMILES centered on the corresponding atom and returns the constructed Root-align SMILES $S_R$({\tt +R}).

\begin{algorithm}[!htb]
    \caption{Root-align SMILES Construction}
    \label{algorithm3}
    \textbf{Input}:  atom-mapped product SMILES $S_P$ and atom-mapped reactants SMILES $S_R$\\
    \textbf{Output}: reactants R-SMILES $S_R$({\tt +R})
    \begin{algorithmic}[1]
        \State Get atom-mapped number list $am(S_P)$ of $S_P$ by Regular Expression Matching.
        \For{$pNum \in am(S_P)$}
            \For{$r_i \in \operatorname{SmiTokenizer}(S_R)$}
                \If {$r_i$ is atom-mapped tokens and $am(r_i)==pNum$}
                    \State Make $r_i$ as root-atom of $S_R$ and get reactants R-SMILES $S_R$({\tt +R})
                    \State \textbf{return} $S_R$({\tt +R})
                \EndIf
            \EndFor
        \EndFor
        \State \textbf{return} $S_R$({\tt +R})
    \end{algorithmic}
\end{algorithm}

\subsection{SMILES Alignment Map Construction}
The ground truth SMILES Alignment Map $SAM$ between the product SMILES $S_P$ and the reactants SMILES $S_R$ is computed as Algorithm~\ref{algorithm2}. 

The algorithm first initializes an empty $SAM$ to store the correspondence between $S_R$ and $S_P$. For each atom-mapped token in $S_R$, the algorithm looks for a corresponding token with the same atom-mapped number in $S_P$. Once this correspondence is found, the algorithm adds the pair of tokens to the $SAM$. This process will traverse forward and backward until there are no more matching tokens. Finally, the algorithm returns the constructed SMILES Alignment Map $SAM$.

\begin{algorithm}[!htb]
    \caption{SMILES Alignment Map Construction}
    \label{algorithm2}
    \textbf{Input}: atom-mapped product SMILES $S_P$ and atom-mapped reactants SMILES $S_R$\\
    \textbf{Output}: SMILES Alignment Map $SAM$
    \begin{algorithmic}[1]
       \State Initialize SMILES Alignment Map $SAM$.
       \For{}{$r_i \in \operatorname{SmiTokenizer}(S_R)$}
            \If{$r_i$ is not visited and $r_i$ is an atom-mapped token}
                \State Find the token $p_j \in \operatorname{SmiTokenizer}(S_p)$ with the same atom-map number as $r_i$: ${am}(p_j) == {am}(r_i)$.
                \While{$r_i == p_j$ or ${am}(p_j) == {am}(r_i)$}
                    \State Add token alignment pair \{$r_i:p_j$\} into $SAM$.
                    \State $i=i+1$ and $j=j+1$.
                \EndWhile
            \EndIf
       \EndFor
       \For{\{$r_i:p_j$\} $\in SAM$}
            \State $i=i-1$ and $j=j-1$.
            \While{$r_i == p_j$ and $r_i$ is not an atom token}
                \State Add token alignment pair \{$r_i:p_j$\} into $SAM$.
                \State $i=i-1$ and $j=j-1$
            \EndWhile
       \EndFor
       \State \textbf{return} $SAM$
    \end{algorithmic}
\end{algorithm}

\section{Hyperparameters Setting}

\label{sec:appe}

\begin{table}[H]
\centering
\label{app:table4}
\caption{ConfRetro hyperparameters setup during training}
\begin{tabular}{lc}
\toprule
\textbf{Hyperparameter}                     &\textbf{value}  \\ \midrule
Layers                                      & 6              \\
Attention heads                             & 8              \\
Embedding dim                               & 512            \\
FFN hidden dim                              & 2048           \\ 
Gaussian kernel channels                    & 512            \\ 
FFN dropout                                 & 0.1            \\
Attention dropout                           & 0.1            \\
Embedding dropout                           & 0.1            \\ \midrule
Coefficient for KL-loss ($\alpha$)          & 0.5            \\
Activation function                         & ReLU           \\ \midrule
Vocabulary size                             & 84             \\ 
Batch size                                  & 16             \\
Max training epoch                          & 500            \\
Beam Search Size                            & 10             \\ \midrule
Warmup steps                                & 8000           \\
Weight decay                                & 1e-3           \\
Learning Rate Scheduler                     & Noam           \\
Learning rate decay                         & 0.001          \\
Peak learning rate                          & 23.8           \\
Adams $\epsilon$                            & 0.5            \\
Adams ($\beta_1, \beta_2$)                  & (0.9, 0.98)    \\
\hline
\end{tabular}

\end{table}

\section{Visualization}

\subsection{Self-Attention Map}

\begin{figure}[!htb]
  \centering
  \includegraphics[width=0.47\linewidth]{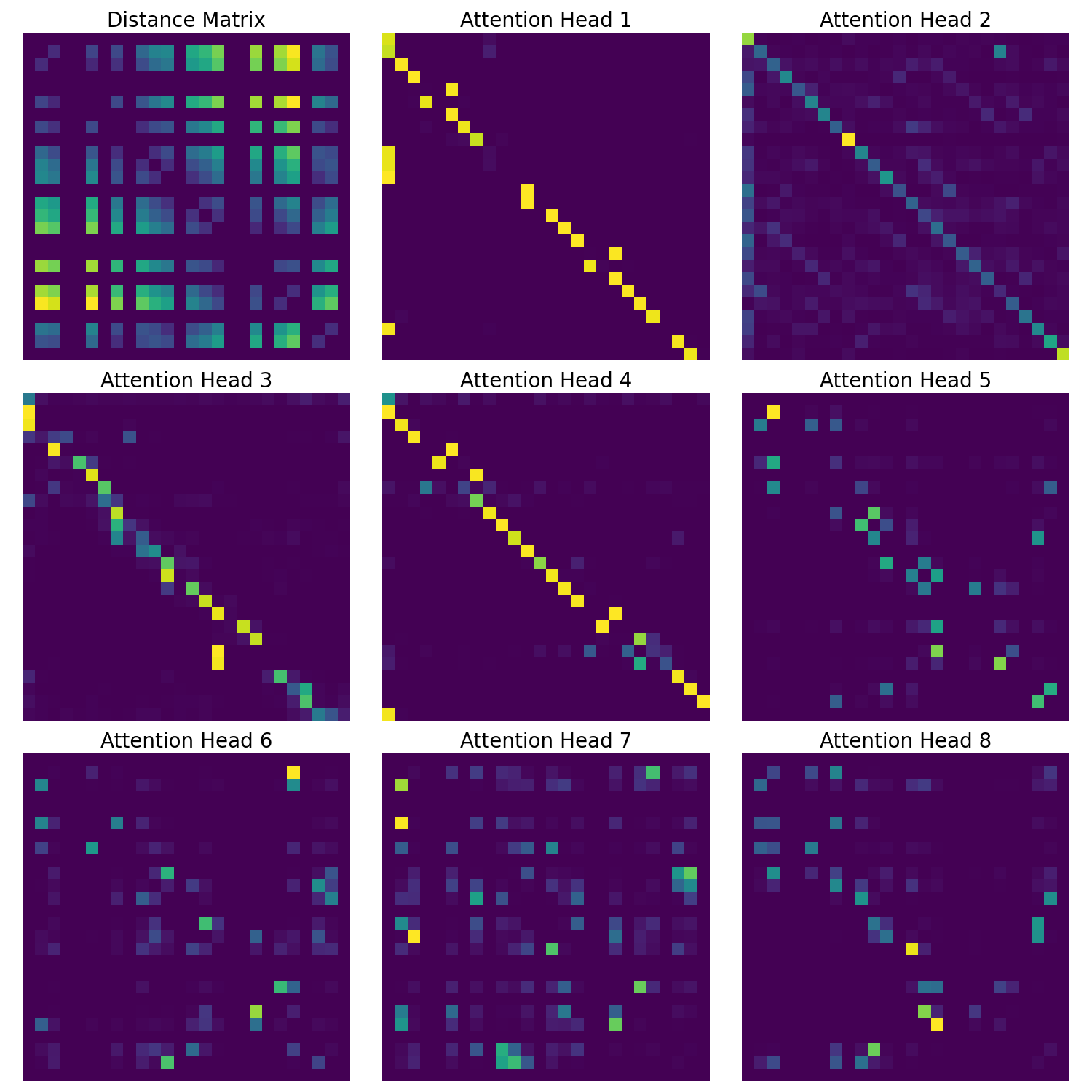}
  \caption{Visualization on self-attention map of multi heads independently.} 
  \label{fig6}
\end{figure}

\begin{figure}[!htb]
  \centering
  \includegraphics[width=0.5\linewidth]{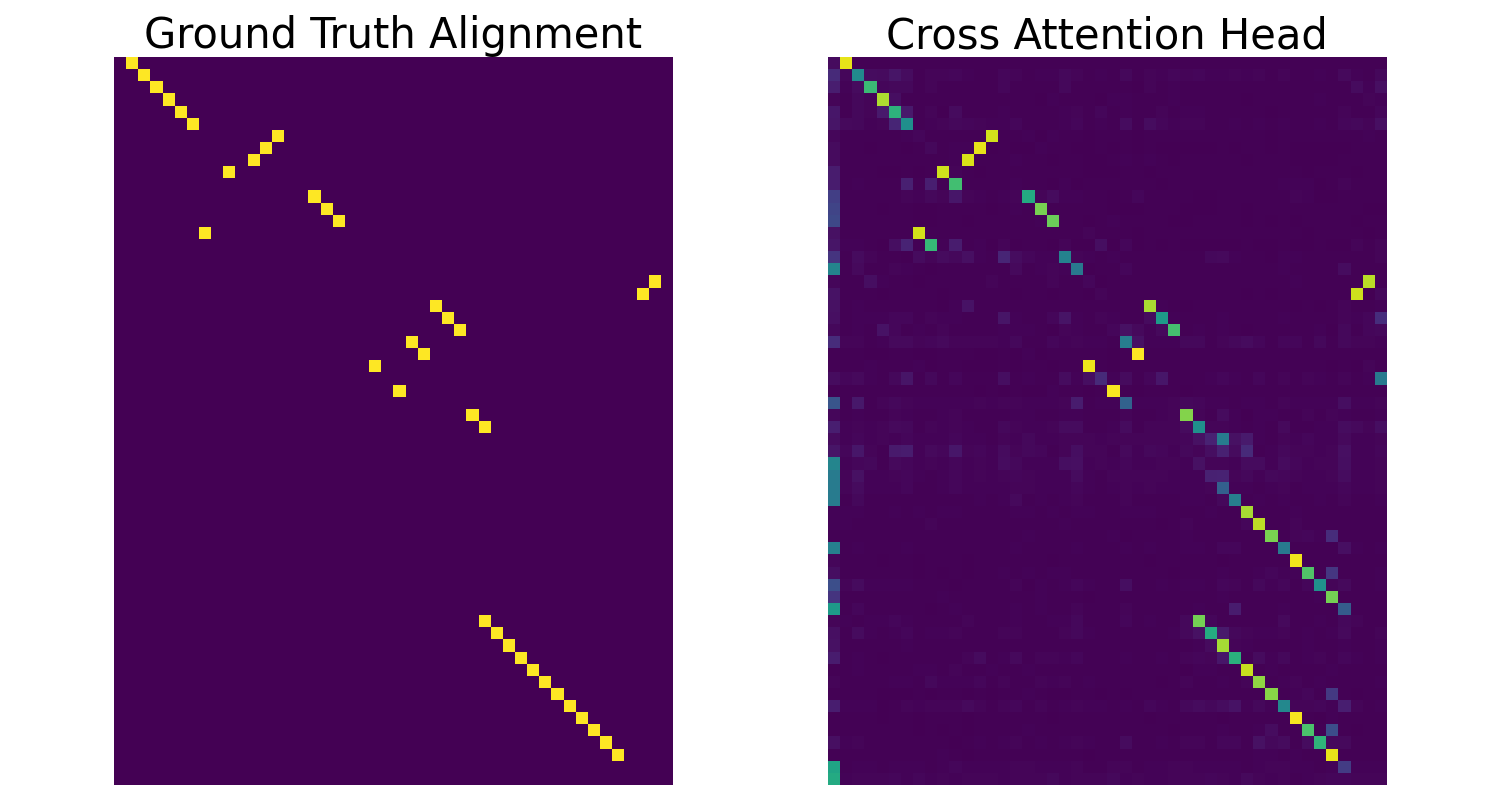}
  \caption{Visualization on cross-attention map on SMILES Alignment.} 
  \label{fig7}
\end{figure}

For better interpretability, we visualize the molecular self-attention maps and distance pair matrices as shown in Figure~\ref{fig6}. It is evident that the sum of the spatial attention heads (Attention Heads 5, 6, 7, 8 in Figure~\ref{fig7}) closely resembles the distance matrix, suggesting that spatial information has been integrated into the self-attention computation. However, when we check these heads independently, different patterns emerge. For example, the normal attention head weights are asymmetric, focusing more on self-token interactions, whereas the distance matrix tends to interact with other tokens. Additionally, long-range interactions can be captured through spatial attention (such as in Attention Heads 7 and 8).

\subsection{Cross-Attention Map}
SMILES Alignment between product and reactants and Cross Attention Score on first head, respectively. The left most figure is the ground truth alignment matrix. It is easy to find that cross-attention captures the corresponding information of the SMILES Alignment Map very well and assists the model generation process in the Decoder stage, which also promotes the final excellent performance of our model.

\section{Case Study}

\label{sec:case}

Here we show \textbf{statistic analysis} and \textbf{more illustrations} in Case Study on \textbf{Intricate Molecules}.

\begin{table}[!htb]
\centering
\label{table7}
\caption{Performance comparison of ConfRetro, NAG2G, and Vanilla Transformer across four types of molecular structures on USPTO-50K (reaction class unknown). Best per column in \textbf{bold}.}
\begin{tabular}{llccccccccc}
\toprule
\multirow{2}{*}{\textbf{Type}} & \multicolumn{1}{c}{\multirow{2}{*}{\textbf{Model}}} & \multicolumn{4}{c}{\textbf{Top-k Accuracy (\%)}} & & \multicolumn{4}{c}{\textbf{Top-k Validity (\%)}} \\ 
& & 1 & 3 & 5 & 10 & & 1 & 3 & 5 & 10 \\ 
\midrule
\multirow{3}{*}{\textbf{Polychiral}} 
& Trans.  & 38.5 & 59.0 & 67.5 & 73.9 & & 97.3 & 97.3 & 96.2 & 92.6 \\ 
& NAG2G   & 41.9 & 63.9 & 71.2 & 77.6 & & \textbf{99.6} & \textbf{98.3} & 96.5 & 91.5 \\ 
& \textbf{ConfRetro} & \textbf{45.7} & \textbf{66.2} & \textbf{75.7} & \textbf{83.7} & & 98.6 & 98.1 & \textbf{97.4} & \textbf{96.2} \\ 
\midrule
\multirow{3}{*}{\textbf{Heteroaromatic}} 
& Trans.  & 51.2 & 71.9 & 79.0 & 84.7 & & 98.2 & 97.6 & 96.4 & 91.1 \\ 
& NAG2G   & 56.2 & 77.5 & 83.9 & 90.2 & & \textbf{99.7} & 98.7 & 97.2 & 92.9 \\ 
& \textbf{ConfRetro} & \textbf{56.8} & \textbf{79.3} & \textbf{86.1} & \textbf{91.8} & & 99.3 & \textbf{98.9} & \textbf{98.7} & \textbf{96.1} \\ 
\midrule
\multirow{3}{*}{\textbf{Fused / Bridged Rings}} 
& Trans.  & 50.8 & 71.1 & 77.8 & 83.3 & & 99.2 & 98.1 & 97.1 & 91.6 \\ 
& NAG2G   & 55.6 & 76.1 & 82.5 & 89.3 & & \textbf{99.5} & \textbf{98.5} & 96.8 & 92.3 \\ 
& \textbf{ConfRetro} & \textbf{56.8} & \textbf{78.0} & \textbf{85.1} & \textbf{91.1} & & 99.4 & 98.4 & \textbf{97.9} & \textbf{95.8} \\ 
\midrule
\multirow{3}{*}{\textbf{Complex}} 
& Trans.  & 49.5 & 70.3 & 77.5 & 83.2 & & 98.3 & 97.6 & 96.3 & 92.0 \\ 
& NAG2G   & 54.7 & 75.4 & 81.6 & 88.5 & & \textbf{99.6} & 98.6 & 96.9 & 92.3 \\ 
& \textbf{ConfRetro} & \textbf{55.7} & \textbf{77.4} & \textbf{84.5} & \textbf{90.4} & & 99.4 & \textbf{98.8} & \textbf{98.1} & \textbf{95.5} \\ 
\hline
\end{tabular}

\end{table}

\begin{figure}[!htb]
  \centering
  \includegraphics[width=0.55\linewidth]{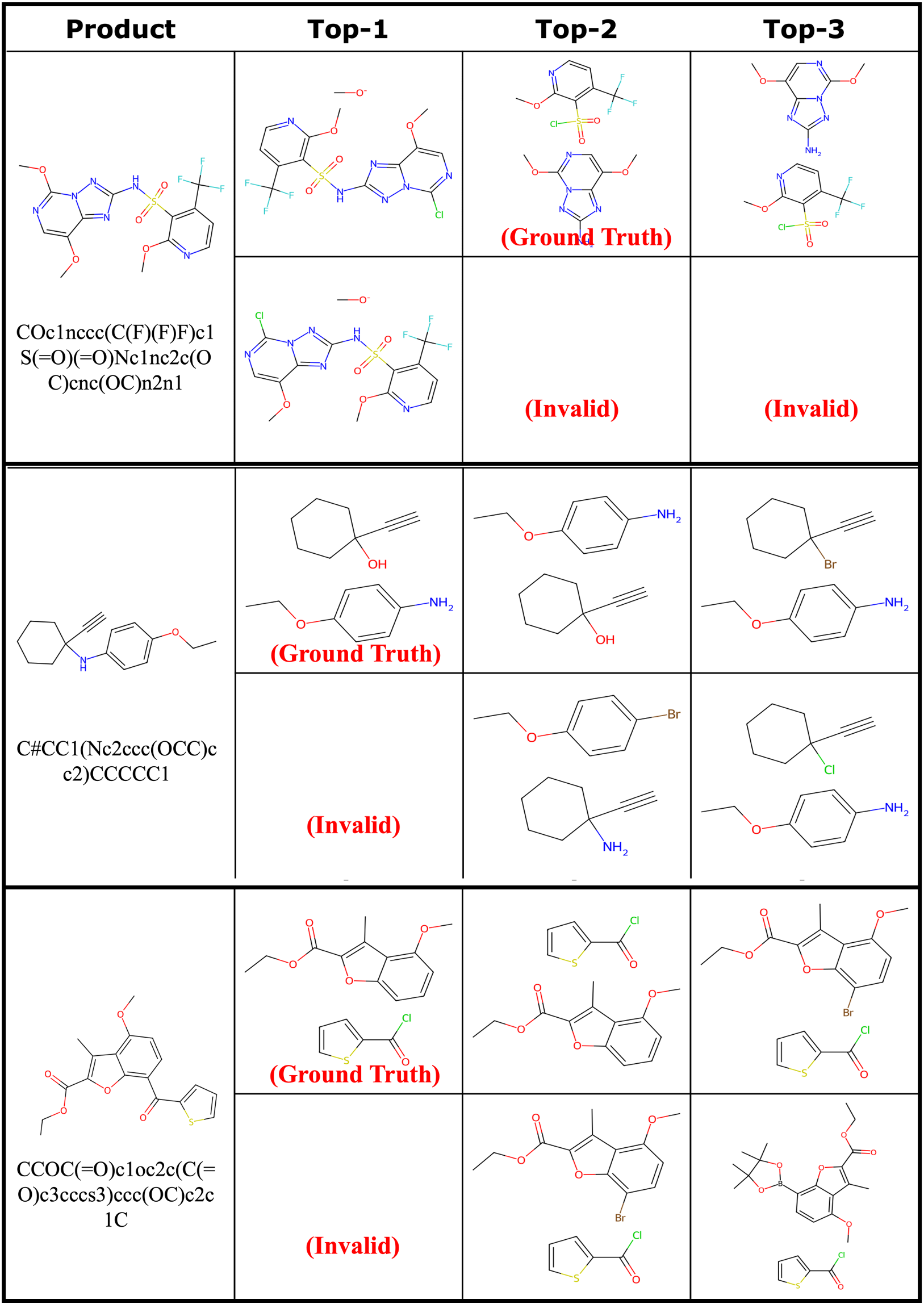}
  \caption{More Case. Each subfigure shows predictions of different methods: \textbf{ConfRetro (Top)}, \textbf{Transformer (Bottom)}.} 
\end{figure}



\section{Synthesis route}

\label{sec:synroute}

\begin{figure}[b]
  \centering
  \includegraphics[width=0.63\linewidth]{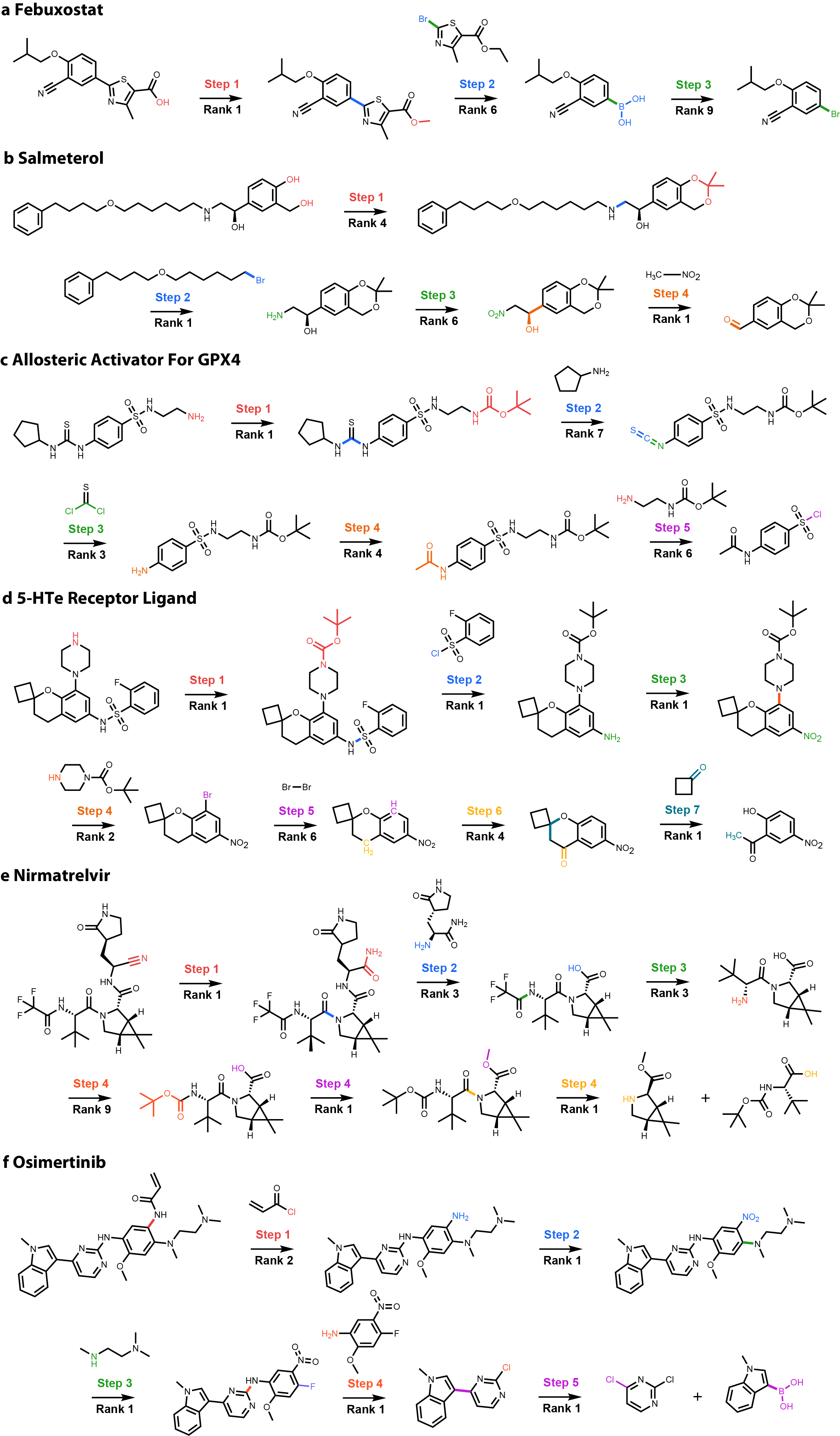}
  \caption{Multistep retrosynthesis predictions by our method. (a) Febuxostat. (b) Salmeterol. (c) Allosteric Activator for GPX4. (d) 5-HTe Receptor Ligand.}
  \label{fig9}
\end{figure}

By applying our product-reactant variant recursively, we verify our method with several multistep retrosynthesis examples reported in the literature, including febuxostat~\citep{febuxostat}, salmeterol~\citep{salmeterol}, an allosteric activator for GPX$_4$~\citep{gpx4}, a 5-HT$_6$ receptor ligand~\citep{5hte}, Nirmatrelvir~\citep{nirmatrelvir} and Osimertinib~\citep{osimertinib}. As shown in Figure~\ref{fig9}
, our method successfully predicts the complete synthetic pathway for these examples.

\section{Robustness and Stability Analysis}

\subsection{Multi-Seed Training Stability}
\label{sec:multiseed}

To assess the reproducibility of reported results, we trained ConfRetro with five independent random seeds $\{10, 1, 42, 123, 2024\}$ under the same protocol (USPTO-50K, reaction class unknown, $n_{\text{aug}}$=20, beam~10). Table~\ref{tab:seeds} reports per-seed results together with the mean and standard deviation.

\begin{table}[!htb]
\centering
\caption{Top-$k$ accuracy across five random seeds (USPTO-50K, reaction class unknown).}
\label{tab:seeds}
\begin{tabular}{lcccc}
\toprule
\textbf{Seed} & \textbf{Top-1} & \textbf{Top-3} & \textbf{Top-5} & \textbf{Top-10} \\ \midrule
10 & 56.2 & 79.3 & 86.2 & 91.7 \\
1 & 56.5 & 79.8 & 86.7 & 92.0 \\
42 & 55.9 & 78.8 & 85.7 & 91.5 \\
123 & 56.3 & 79.5 & 86.4 & 91.7 \\
2024 & 56.1 & 79.1 & 86.0 & 91.6 \\ \midrule
\textbf{Mean $\pm$ Std} & \textbf{56.2$\pm$0.2} & \textbf{79.3$\pm$0.4} & \textbf{86.2$\pm$0.4} & \textbf{91.7$\pm$0.2} \\
\bottomrule
\end{tabular}
\end{table}

The Top-1 standard deviation is $\pm$0.2\%, confirming that reported results are stable across training runs.

\subsection{Conformer-Seed Robustness}
\label{sec:confseed}

To verify that performance is insensitive to the specific RDKit conformer used at test time, we evaluated a trained ConfRetro model on five different RDKit embedding seeds. Table~\ref{tab:confseed} reports per-seed results together with the mean and standard deviation. The Top-1 standard deviation is $\pm$0.1\%, demonstrating that ConfRetro's performance is essentially invariant to which conformer is used. A five-conformer reciprocal-rank fusion ensemble achieves $56.3/79.4/86.3/91.8$, on par with a single conformer, confirming that a cheap single deterministic conformer suffices.

\begin{table}[!htb]
\centering
\caption{Top-$k$ accuracy across five RDKit conformer-embedding seeds (USPTO-50K, reaction class unknown). Seed 10 is the default conformer used throughout the paper.}
\label{tab:confseed}
\begin{tabular}{lcccc}
\toprule
\textbf{Seed} & \textbf{Top-1} & \textbf{Top-3} & \textbf{Top-5} & \textbf{Top-10} \\ \midrule
10 & 56.2 & 79.3 & 86.2 & 91.7 \\
7 & 56.1 & 79.2 & 86.1 & 91.4 \\
23 & 56.3 & 79.4 & 86.3 & 91.9 \\
88 & 56.2 & 79.3 & 86.2 & 91.7 \\
256 & 56.2 & 79.3 & 86.2 & 91.8 \\ \midrule
\textbf{Mean $\pm$ Std} & \textbf{56.2$\pm$0.1} & \textbf{79.3$\pm$0.1} & \textbf{86.2$\pm$0.1} & \textbf{91.7$\pm$0.2} \\
\bottomrule
\end{tabular}
\end{table}

\subsection{Stereochemistry-Stratified Evaluation}
\label{sec:stereo}

To verify that ConfRetro's gains are not confounded by stereochemical annotation quality, we stratified the USPTO-50K test reactions by the stereochemistry of the product into three groups: \emph{defined} (the product has at least one stereocentre whose configuration is explicitly assigned in the SMILES, e.g.\ {\tt @}/{\tt @@} or E/Z; $n$=870, 17.4\%); \emph{unspecified} (the product contains at least one stereocentre, but its configuration is left unassigned in the SMILES, i.e.\ a potential chiral atom without an {\tt @}/{\tt @@} label; $n$=940, 18.8\%); and \emph{none} (the product has no stereocentre at all; $n$=3,180, 63.7\%). The \emph{defined} and \emph{unspecified} groups thus differ only in whether the existing stereocentres carry an explicit configuration. In each stratum we compare the standard ConfRetro, which uses the real RDKit conformer, against a \emph{random-conformer} control: an otherwise identical model whose 3D atomic coordinates are replaced by random noise (the same control as ``w/ random conformer'' in main-text Table~5), so that the gap isolates the contribution of genuine 3D geometry. Table~\ref{tab:stereo} reports the results per stratum.

\begin{table}[!htb]
\centering
\caption{Stereochemistry-stratified Top-$k$ accuracy: ConfRetro (real conformer) vs.\ the random-conformer control (3D coordinates replaced by random noise).}
\label{tab:stereo}
\begin{tabular}{llcccc}
\toprule
\textbf{Stratum} & \textbf{Model} & \textbf{Top-1} & \textbf{Top-3} & \textbf{Top-5} & \textbf{Top-10} \\ \midrule
\multirow{2}{*}{All} & ConfRetro & 56.2 & 79.3 & 86.2 & 91.7 \\
 & Random & 53.4 & 77.7 & 84.8 & 90.7 \\ \midrule
\multirow{2}{*}{Defined} & ConfRetro & 51.7 & 71.0 & 80.0 & 86.8 \\
 & Random & 49.4 & 69.7 & 76.7 & 84.0 \\ \midrule
\multirow{2}{*}{Unspecified} & ConfRetro & 52.0 & 76.5 & 83.8 & 89.4 \\
 & Random & 48.5 & 73.4 & 81.6 & 88.2 \\ \midrule
\multirow{2}{*}{None} & ConfRetro & 58.6 & 82.5 & 88.7 & 93.8 \\
 & Random & 56.0 & 81.2 & 88.0 & 93.3 \\
\bottomrule
\end{tabular}
\end{table}

ConfRetro outperforms the random-conformer control across all strata, with the largest Top-1 gain on the \emph{unspecified} stratum ($+$3.5), showing that the model benefits from 3D geometry even when stereo annotations are absent or ambiguous.

\subsection{Checkpoint-Count Ablation}
\label{sec:ckpt}

Table~\ref{tab:ckpt} reports the effect of the number of averaged checkpoints. Top-1 accuracy rises from 55.4 (3 checkpoints) to 56.2 (7 checkpoints) then plateaus; we therefore selected $k$=7 as the best and most efficient choice.

\begin{table}[!htb]
\centering
\caption{Effect of checkpoint averaging count on USPTO-50K (reaction class unknown).}
\label{tab:ckpt}
\begin{tabular}{lcccc}
\toprule
\textbf{Checkpoints} & \textbf{Top-1} & \textbf{Top-3} & \textbf{Top-5} & \textbf{Top-10} \\ \midrule
3 & 55.4 & 78.5 & 85.4 & 91.3 \\
5 & 55.9 & 78.9 & 85.8 & 91.5 \\
\textbf{7 (selected)} & \textbf{56.2} & 79.3 & 86.2 & 91.7 \\
9 & 56.1 & 79.2 & 86.1 & 91.6 \\
\bottomrule
\end{tabular}
\end{table}

\subsection{Statistical Significance Tests}
\label{sec:sig}

To confirm that the accuracy gaps are real rather than test-set noise, we run two analyses on the 4,990 USPTO-50K test reactions (reaction class unknown). \emph{(i) Bootstrap confidence intervals (CIs).} We resample the test reactions with replacement $B$=10{,}000 times and take the 2.5th and 97.5th percentiles of each metric as its 95\% CI (Table~\ref{tab:ci}); a narrow CI means a stable estimate, and two CIs that barely overlap indicate a real gap. \emph{(ii) McNemar tests.} Since all models are tested on the same reactions, we compare them in a paired manner using the exact two-sided McNemar test, which looks only at the reactions solved by exactly one of the two models at Top-1 and asks whether such disagreements are balanced; a small $p$-value means they are too one-sided to be due to chance (Table~\ref{tab:mcnemar}). The \emph{Diff} column gives the Top-1 accuracy difference (A$-$B) with its bootstrap 95\% CI.

\begin{table}[!htb]
\centering
\caption{Bootstrap 95\% confidence intervals (percentile, $B$=10{,}000) on USPTO-50K.}
\label{tab:ci}
\begin{tabular}{lcccc}
\toprule
\textbf{Model} & \textbf{Top-1} & \textbf{Top-3} & \textbf{Top-5} & \textbf{Top-10} \\ \midrule
ConfRetro & 56.2 [54.8, 57.5] & 79.3 [78.2, 80.4] & 86.2 [85.2, 87.2] & 91.7 [90.9, 92.5] \\
w/o bond-type  & 56.0 [54.5, 57.4] & 79.2 [78.1, 80.4] & 86.5 [85.5, 87.4] & 91.8 [91.0, 92.6] \\
w/ random conformer   & 53.4 [52.1, 54.8] & 77.7 [76.5, 78.9] & 84.8 [83.8, 85.8] & 90.7 [89.9, 91.5] \\
w/o 3D distance  & 53.0 [51.5, 54.3] & 77.4 [76.3, 78.6] & 84.5 [83.5, 85.5] & 90.5 [89.6, 91.3] \\
Retroformer        & 53.6 [52.3, 55.0] & 73.7 [72.4, 74.9] & 80.2 [79.0, 81.3] & 86.1 [85.1, 87.0] \\
\bottomrule
\end{tabular}
\end{table}

\begin{table}[!htb]
\centering
\caption{Paired significance: difference in accuracy and McNemar $p$-value.}
\label{tab:mcnemar}
\begin{tabular}{llcc}
\toprule
\textbf{A} & \textbf{B} & \textbf{Diff (A$-$B)} & \textbf{McNemar $p$} \\ \midrule
ConfRetro & w/ random conformer & $+$2.71 [+1.72, +3.67] & $9.0\!\times\!10^{-8}$ \\ 
ConfRetro & w/o 3D distance & $+$3.19 [+2.18, +4.19] & $4.3\!\times\!10^{-10}$ \\
w/o bond-type & w/o 3D distance & $+$3.01 [+2.04, +3.97] & $1.6\!\times\!10^{-9}$ \\
ConfRetro & Retroformer (Top-1) & $+$2.55 & $1.9\!\times\!10^{-5}$ \\
ConfRetro & Retroformer (Top-3) & $+$4.73 & $5.2\!\times\!10^{-20}$ \\
ConfRetro & Retroformer (Top-5) & $+$4.99 & $7.0\!\times\!10^{-27}$ \\
ConfRetro & Retroformer (Top-10) & $+$5.03 & $2.4\!\times\!10^{-37}$ \\
\bottomrule
\end{tabular}
\end{table}

All geometry-related improvements are statistically significant at $p < 10^{-6}$.

\end{document}